\documentclass[10pt,twocolumn,letterpaper]{article}

\usepackage[pagenumbers]{cvpr}
\usepackage[table]{xcolor}
\usepackage{graphicx}
\usepackage{amsmath}
\usepackage{amssymb}
\usepackage{booktabs}
\usepackage{makecell}
\usepackage{array}
\usepackage{wrapfig}
\usepackage{multirow}
\usepackage{bbm}
\usepackage{float}
\usepackage{graphicx}
\usepackage{perpage}
\usepackage{enumitem}
\usepackage{algpseudocode}
\usepackage[ruled,vlined]{algorithm2e}
\usepackage{extarrows}
\usepackage{amssymb}
\usepackage{booktabs}
\usepackage{marvosym}
\usepackage{pifont}
\usepackage{gensymb}

\usepackage[pagebackref=true,breaklinks=true,letterpaper=true,colorlinks,urlcolor=black,bookmarks=false]{hyperref}

\newlength\savedwidth
\newcommand\whline{\noalign{\global\savedwidth\arrayrulewidth\global\arrayrulewidth 0.8pt}\hline\noalign{\global\arrayrulewidth\savedwidth}}

\newcommand{\green}[1]{\textcolor[RGB]{96,177,87}{#1}}

\newcommand{\yes}{\ding{51}}
\newcommand{\no}{\ding{55}}

\def\ie{\emph{i.e.}}
\def\eg{\emph{e.g.}}

\def\etal{{\em et al.~}}
\def\model{InternImage}

\definecolor{mygray}{gray}{.92}

\newcommand\blfootnote[1]{%
\begingroup
\renewcommand\thefootnote{}\footnote{#1}%
\addtocounter{footnote}{-1}%
\endgroup
}

\begin{document}

\title{\model: Exploring Large-Scale Vision Foundation Models with \\Deformable Convolutions}

\author{
    Wenhai Wang$^{1*}$, 
    Jifeng Dai$^{2,1*}$,
    Zhe Chen$^{3,1*}$,
    Zhenhang Huang$^{1*}$,
    Zhiqi Li$^{3,1*}$,
    Xizhou Zhu$^{4*}$,\\
    Xiaowei Hu$^{1}$,
    Tong Lu$^{3}$,
    Lewei Lu$^{4}$,
    Hongsheng Li$^{5}$,
    Xiaogang Wang$^{4,5}$,
    Yu Qiao$^{1}$\textsuperscript{\Letter}\\
    $^1$ Shanghai AI Laboratory~~~
    $^2$Tsinghua University~~~\\
    $^3$Nanjing University~~~
    $^4$SenseTime Research~~~
    $^5$The Chinese University of Hong Kong\\
    {\small \url{https://github.com/OpenGVLab/InternImage}}
}

\maketitle

\begin{abstract}
  Compared to the great progress of large-scale vision transformers (ViTs) in recent years, large-scale models based on convolutional neural networks (CNNs) are still in an early state. This work presents a new large-scale CNN-based foundation model, termed InternImage, which can obtain the gain from increasing parameters and training data like ViTs. Different from the recent CNNs that focus on large dense kernels, InternImage takes deformable convolution as the core operator, so that our model not only has the large effective receptive field required for downstream tasks such as detection and segmentation, but also has the adaptive spatial aggregation conditioned by input and task information. As a result, the proposed InternImage reduces the strict inductive bias of traditional CNNs and makes it possible to learn stronger and more robust patterns with large-scale parameters from massive data like ViTs. The effectiveness of our model is proven on challenging benchmarks including ImageNet, COCO, and ADE20K. It is worth mentioning that InternImage-H achieved a new record 65.4 mAP on COCO test-dev and 62.9 mIoU on ADE20K, outperforming current leading CNNs and ViTs.

\end{abstract}

\section{Introduction}

\begin{figure}[t]
    \centering
    \includegraphics[width=1.0\columnwidth]{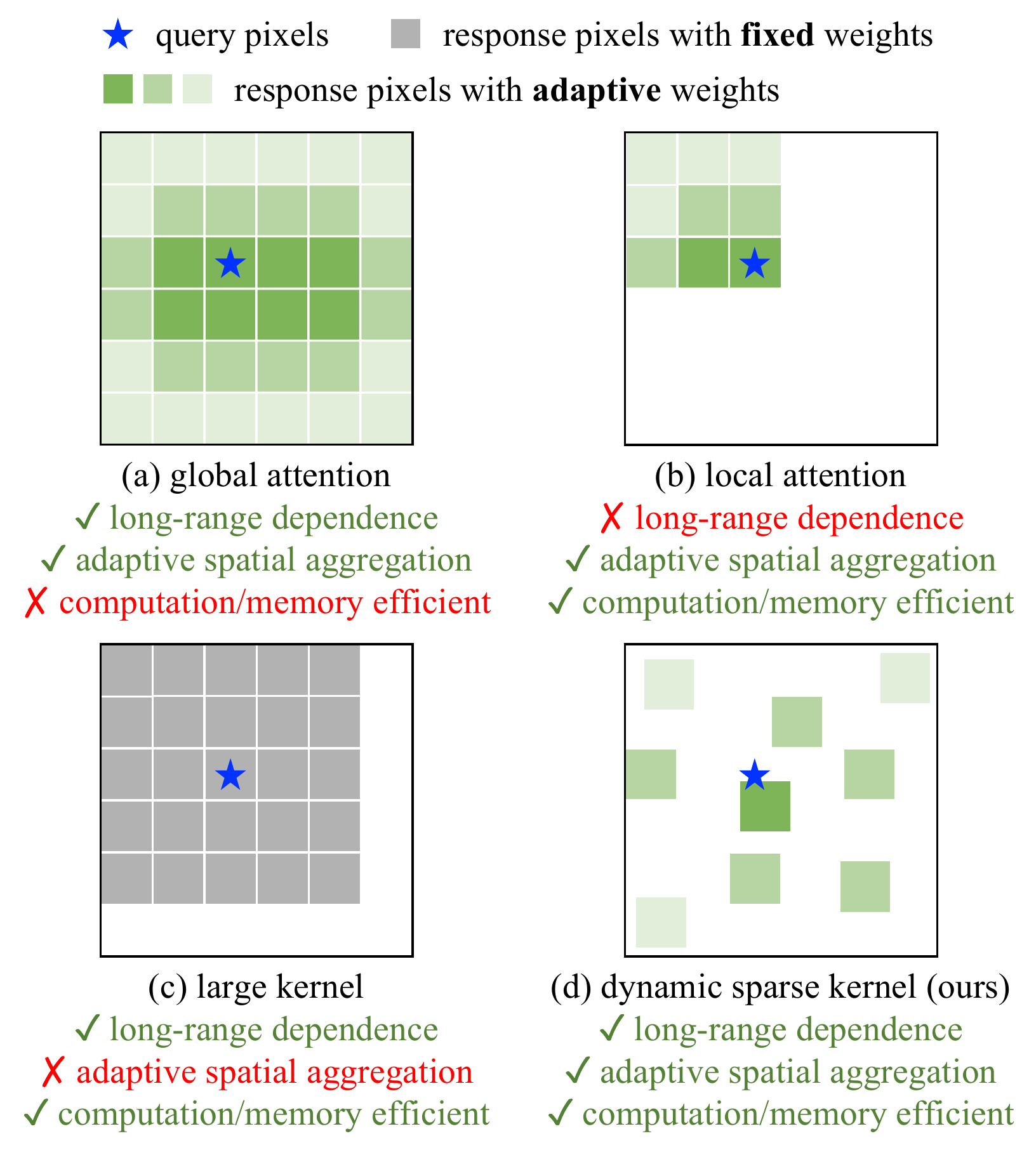}
    \caption{
    \textbf{Comparisons of different core operators.} (a) shows the global aggregation of multi-head self-attention (MHSA)~\cite{vaswani2017attention}, whose computational and memory costs are expensive in downstream tasks that require high-resolution inputs. (b) limits the range of MHSA into a local window~\cite{liu2021swin} to reduce the cost. (c) is a depth-wise convolution with very large kernels to model long-range dependencies. (d) is a deformable convolution, which shares similar favorable properties with MHSA and is efficient enough for large-scale models.
    We start from it to build a large-scale CNN.
    }
    \label{fig:coreop}
\end{figure}

\blfootnote{* equal contribution, \Letter\ corresponding author (qiaoyu@pjlab.org.cn)}
With the remarkable success of transformers in large-scale language models~\cite{shoeybi2019megatron,radford2019gpt2,raffel2020t5,brown2020gpt3,chowdhery2022palm,fedus2022switch},
vision transformers (ViTs)~\cite{dosovitskiy2020image,liu2021swin,wang2021pyramid,wang2021pvtv2,dong2021cswin,wu2021cvt,ali2021xcit,han2021transformer} have also swept the computer vision field and are becoming the primary choice for the research and practice of large-scale vision foundation models.
Some pioneers~\cite{liu2021swinv2,wang2022beit3,riquelme2021vmoe,zhai2022scalingvit,dai2021coatnet} have made attempts to extend ViTs to very large models with over a billion parameters,
beating convolutional neural networks (CNNs) and significantly pushing the performance bound for a wide range of computer vision tasks, including basic classification, detection, and segmentation.
While these results suggest that CNNs are inferior to ViTs in the era of massive parameters and data, 
we argue that \emph{CNN-based foundation models can also achieve comparable or even better performance than ViTs when equipped with similar operator-/architecture-level designs, scaling-up parameters, and massive data}.

To bridge the gap between CNNs and ViTs, we first summarize their differences from two aspects:
(1) From the operator level~\cite{dosovitskiy2020image,liu2022convnet,ding2022replknet}, the multi-head self-attention (MHSA) of ViTs has long-range dependencies and adaptive spatial aggregation (see Fig. \ref{fig:coreop}(a)).
Benefiting from the flexible MHSA,
ViTs can learn more powerful and robust representations than CNNs from massive data.
(2) From the architecture view~\cite{dosovitskiy2020image,ding2022replknet,yu2022metaformer}, besides MHSA, ViTs contain a series of advanced components that are not included in standard CNNs, such as Layer Normalization (LN)~\cite{ba2016layernorm}, feed-forward network (FFN)~\cite{vaswani2017attention}, GELU~\cite{hendrycks2016gelu}, etc.
Although recent works~\cite{liu2022convnet,ding2022replknet} have made meaningful attempts to introduce long-range dependencies into CNNs by using dense convolutions with very large kernels (\eg, 31$\times$31) as shown in Fig. \ref{fig:coreop} (c),
there is still a considerable gap with the state-of-the-art large-scale ViTs~\cite{zhai2022scalingvit,dai2021coatnet,liu2021swinv2,riquelme2021vmoe,wei2022fdswin} in terms of performance and model scale.

In this work, we concentrate on designing a CNN-based foundation model that can efficiently extend to large-scale parameters and data.
Specifically, we start with a flexible convolution variant---deformable convolution (DCN)~\cite{dai2017dcnv1,zhu2019dcnv2}.
By combining it with a series of tailored block-level and architecture-level designs similar to transformers, we design a brand-new convolutional backbone network, termed \emph{\model}.
As shown in Fig. \ref{fig:coreop}, different from recently improved CNNs with very large kernels such as 31$\times$31~\cite{ding2022replknet},
the core operator of \model\ is a dynamic sparse convolution with a common window size of 3$\times$3, 
(1) whose sampling offsets are flexible to dynamically learn appropriate receptive fields (can be long- or short-range) from given data;
(2) the sampling offsets and modulation scalars are 
adaptively adjusted according to the input data, which can achieve adaptive spatial aggregation like ViTs, reducing the over-inductive bias of regular convolutions;
and (3) the convolution window is a common 3$\times$3, avoiding the optimization problems and expensive costs caused by large dense kernels~\cite{ding2022replknet,liu2022slak}.

With the aforementioned designs, the proposed \model\
can efficiently scale to large parameter sizes and learn stronger representations from large-scale training data,
achieving comparable or even better performance to large-scale ViTs~\cite{liu2021swin,wang2021pvtv2,zhai2022scaling} on a wide range of vision tasks.
In summary, our main contributions are as follows:

(1) 
We present a new large-scale CNN-based foundation model---\model.
To our best knowledge, it is the first CNN that effectively scales to over 1 billion parameters and 400 million training images and achieves comparable or even better performance than state-of-the-art ViTs, showing that convolutional models are also a worth-exploring direction for large-scale model research.

(2) 
We successfully scale CNNs to large-scale settings by introducing long-range dependencies and adaptive spatial aggregation using an improved 3$\times$3 DCN operator,
and explore the tailored basic block, stacking rules, and scaling strategies centered on the operator.
These designs make effective use of the operator, enabling our models to obtain the gains from large-scale parameters and data.

\begin{figure}[t]
    \centering
   \includegraphics[width=1.0\columnwidth]{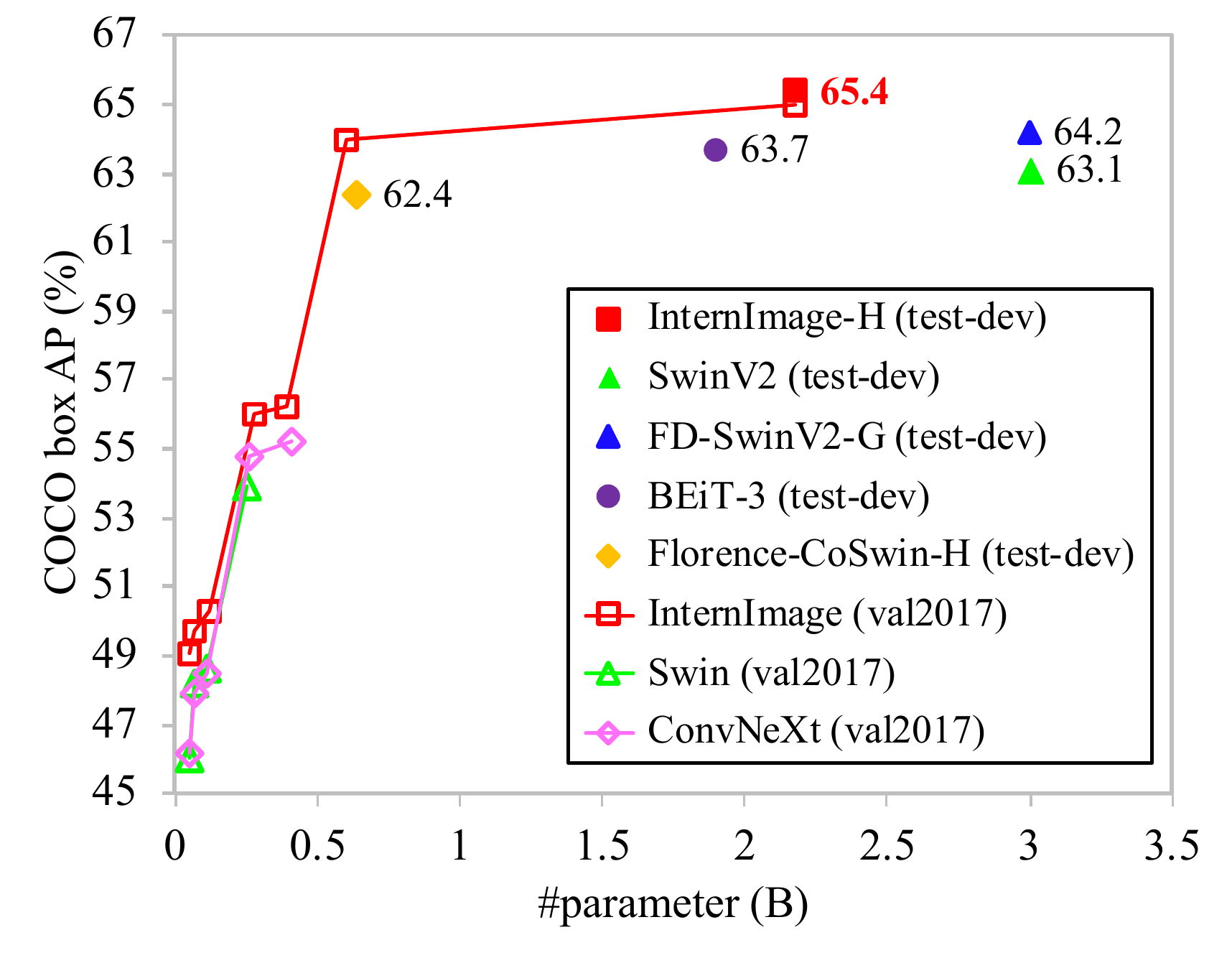}
    \caption{\textbf{Performance comparison on COCO of different backbones.} The proposed \model-H achieves a new record 65.4 box AP on COCO test-dev, significantly outperforming state-of-the-art CNNs and large-scale ViTs.}
    \label{fig:bb_cmp}
\end{figure}

(3) We evaluate the proposed model on representative vision tasks including image classification, object detection, instance and semantic segmentation,
and compared it with state-of-the-art CNNs and large-scale ViTs by scaling the model size ranging from 30 million to 1 billion, the data ranging from 1 million to 400 million.
Specifically, our model with different parameter sizes can consistently outperform prior arts on ImageNet~\cite{deng2009imagenet}. 
\model-B achieves 84.9\% top-1 accuracy trained only on the ImageNet-1K dataset, outperforming CNN-based counterparts~\cite{ding2022replknet,liu2022convnet} by at least 1.1 points.
With large-scale parameters (\ie, 1 billion) and training data (\ie, 427 million), the top-1 accuracy of \model-H is further boosted to 89.6\%, which is close to well-engineering ViTs~\cite{liu2021swin,zhai2022scaling} and hybrid-ViTs~\cite{dai2021coatnet}.
In addition, on COCO~\cite{lin2014microsoft}, a challenging downstream benchmark,
our best model \model-H achieves state-of-the-art 65.4\% box mAP with 2.18 billion parameters, 2.3 points higher than SwinV2-G~\cite{liu2021swinv2} (65.4 \vs 63.1) with 27\% fewer parameters as shown in Fig. \ref{fig:bb_cmp}.

\section{Related Work}

\textbf{Vision foundation models.} Convolutional neural networks (CNNs) became the mainstream for visual recognition after the large-scale dataset and computation resources were available.
Straining from AlexNet~\cite{krizhevsky2017imagenet}, lots of deeper and more effective neural network architectures have been proposed, such as VGG~\cite{simonyan2014very}, GoogleNet~\cite{szegedy2015going}, ResNet~\cite{he2016deep}, ResNeXt~\cite{xie2017aggregated}, 
EfficientNet~\cite{tan2019efficientnet,tan2021efficientnetv2}, etc.
In addition to the architectural design, more sophisticated convolution operations such as depth-wise convolution~\cite{howard2017mobilenets} and deformable convolution~\cite{dai2017dcnv1,zhu2019dcnv2} are formulated.
By considering the advanced designs
of transformers, modern CNNs showed promising performance on the vision tasks by discovering better components in macro/micro designs and introducing improved convolutions with long-range dependencies~\cite{liu2022convnet,ding2022scaling,liu2022more,rao2022hornet} or dynamic weights~\cite{han2021connection}.

In recent years, a new line of vision foundation models focuses on transformer-based architecture.
ViT~\cite{dosovitskiy2020image} is the most representative model,
which achieves great success in vision tasks thanks to global receptive fields and dynamic spatial aggregation.
However, global attention in ViT suffers from expensive computational/memory complexity, especially on large feature maps, which limits its application in downstream tasks.
To address this problem,
PVT~\cite{wang2021pyramid,wang2021pvtv2} and Linformer~\cite{wang2020linformer} performed global attention on the downsampled key and value maps, DAT~\cite{xia2022vision} employed deformable attention to sparsely sample information from value maps, while HaloNet~\cite{vaswani2021scaling} and Swin transformer~\cite{liu2021swin} developed local attention mechanisms and used haloing and shift operations to transfer information among adjacent local regions.

\textbf{Large-scale models.} Scaling up models is an important strategy to improve feature representation quality, which has been well-studied in the natural language processing (NLP) domain~\cite{kaplan2020scaling}. 
Inspired by the success in the NLP field, Zhai \etal~\cite{zhai2022scalingvit} first extended ViT to 2 billion parameters.
Liu \etal~\cite{liu2021swinv2} enlarged the hierarchical-structure Swin transformer to a deeper and wider model with 3 billion parameters.
Some researchers developed large-scale hybrid ViTs~\cite{dai2021coatnet,ding2022davit} by combining the advantages of ViTs and CNNs at different levels.
Recently, BEiT-3~\cite{wang2022beit3} further explored stronger representations based on ViT with large-scale parameters using multimodal pre-training.
These methods significantly raise the upper bound of basic vision tasks.
However, research on CNN-based large-scale models has lagged behind transformer-based architectures in terms of the total number of parameters and performance.
Although newly-proposed CNNs~\cite{liu2022convnet,ding2022scaling,liu2022more,rao2022hornet} introduce long-range dependencies by using convolutions with very large kernels or recursive gated kernels, there is still a considerable gap with state-of-the-art ViTs.
In this work, \emph{we aim to develop a CNN-based foundation model that can extend efficiently to a large scale comparable to ViT.}

\section{Proposed Method}
To design a large-scale CNN-based foundation model, we start with a flexible convolution variant, namely deformable convolution v2 (DCNv2)~\cite{zhu2019dcnv2}
and make some tune-ups based on it 
to better suit the requirements of large-scale foundation models.
Then, we build the basic block by combining the tuned convolution operator with advanced block designs used in modern backbones~\cite{liu2021swinv2,zhai2022scalingvit}.
Finally, we explore the stacking and scaling principles of DCN-based blocks to build a large-scale convolutional model that can learn strong representations from massive data.

\subsection{Deformable Convolution v3}
\textbf{Convolution \vs MHSA.} Previous works~\cite{zhu2019empirical,liu2022convnet,ding2022replknet} have extensively discussed the differences between CNNs and ViTs.
Before deciding on the core operator of \model, we first summarize the main differences between regular convolution and MHSA.

(1) \emph{Long-range dependencies}. Although it has long been recognized that models with large effective receptive fields (long-range dependencies) usually perform better on downstream vision tasks~\cite{chen2017deeplab,florian2017rethinking,chen2018encoder},
the de-facto effective receptive field of CNNs~\cite{simonyan2014very,he2016deep} stacked by 3$\times$3 regular convolutions is relatively small. 
Even with very deep models,
the CNN-based model still cannot acquire long-range dependencies like ViTs, which limits its performance.

(2) \emph{Adaptive spatial aggregation}. 
Compared to MHSA whose weights are dynamically conditioned by the input,
regular convolution~\cite{lecun1989backpropagation} is an operator with static weights and strong inductive biases such as 2D locality, neighborhood structure, translation equivalence, etc.
With the highly-inductive properties, models composed by regular convolutions might converge faster and require less training data than ViTs, 
but it also restricts CNNs from learning more general and robust patterns from web-scale data.

\textbf{Revisiting DCNv2.} A straightforward way to bridge the gap between convolution and MHSA is to introduce long-range dependencies and adaptive spatial aggregation into regular convolutions.
Let us start with DCNv2~\cite{zhu2019dcnv2}, which is a general variant of regular convolution.
Given an input $\mathbf{x}\!\in\!\mathbb{R}^{C\times H\times W}$ and current pixel $p_0$, DCNv2 can be formulated as:
\begin{equation}
    \textbf{y}(p_0) = \sum^{K}_{k=1}\mathbf{w}_k  \mathbf{m}_k \mathbf{x}(p_0 + p_k + \Delta p_k),
    \label{eqn:dcnv2}
\end{equation}
where $K$ represents the total number of sampling points, and $k$ enumerates the sampling point. 
$\mathbf{w}_k\!\in\!\mathbb{R}^{C\times C}$ denotes the projection weights of the $k$-th sampling point,
and $\mathbf{m}_k\!\in\!\mathbb{R}$ represents the modulation scalar of the $k$-th sampling point, which is normalized by sigmoid function.
$p_k$ denotes the $k$-th location of the pre-defined grid sampling $\{(-1, -1), (-1, 0), ..., (0, +1), ..., (+1, +1)\}$ as in regular convolutions,
and $\Delta p_k$ is the offset corresponding to the $k$-th grid sampling location.
We see from the equation that (1) for long-range dependencies, the sampling offset $\Delta p_k$ is flexible and able to interact with short- or long-range features; and (2) for adaptive spatial aggregation, both the sampling offset $\Delta p_k$ and modulation scalar $\mathbf{m}_k$ are learnable and conditioned by input $\mathbf{x}$. 
So it can be found that  \emph{DCNv2 shares similar favorable properties with MHSA}, which motivated us to develop large-scale CNN-based foundation models on the basis of this operator.

\textbf{Extending DCNv2 for Vision Foundation Models.}
In common practice, DCNv2 is usually used as an extension to regular convolutions, loading pre-trained weights of regular convolutions and fine-tuning for better performance,
which is not exactly suitable for large-scale vision foundation models that need to be trained from scratch.
In this work, to address this problem, we extend DCNv2 from aspects as follows:

\begin{figure}[t]
    \centering
    \includegraphics[width=1.0\columnwidth]{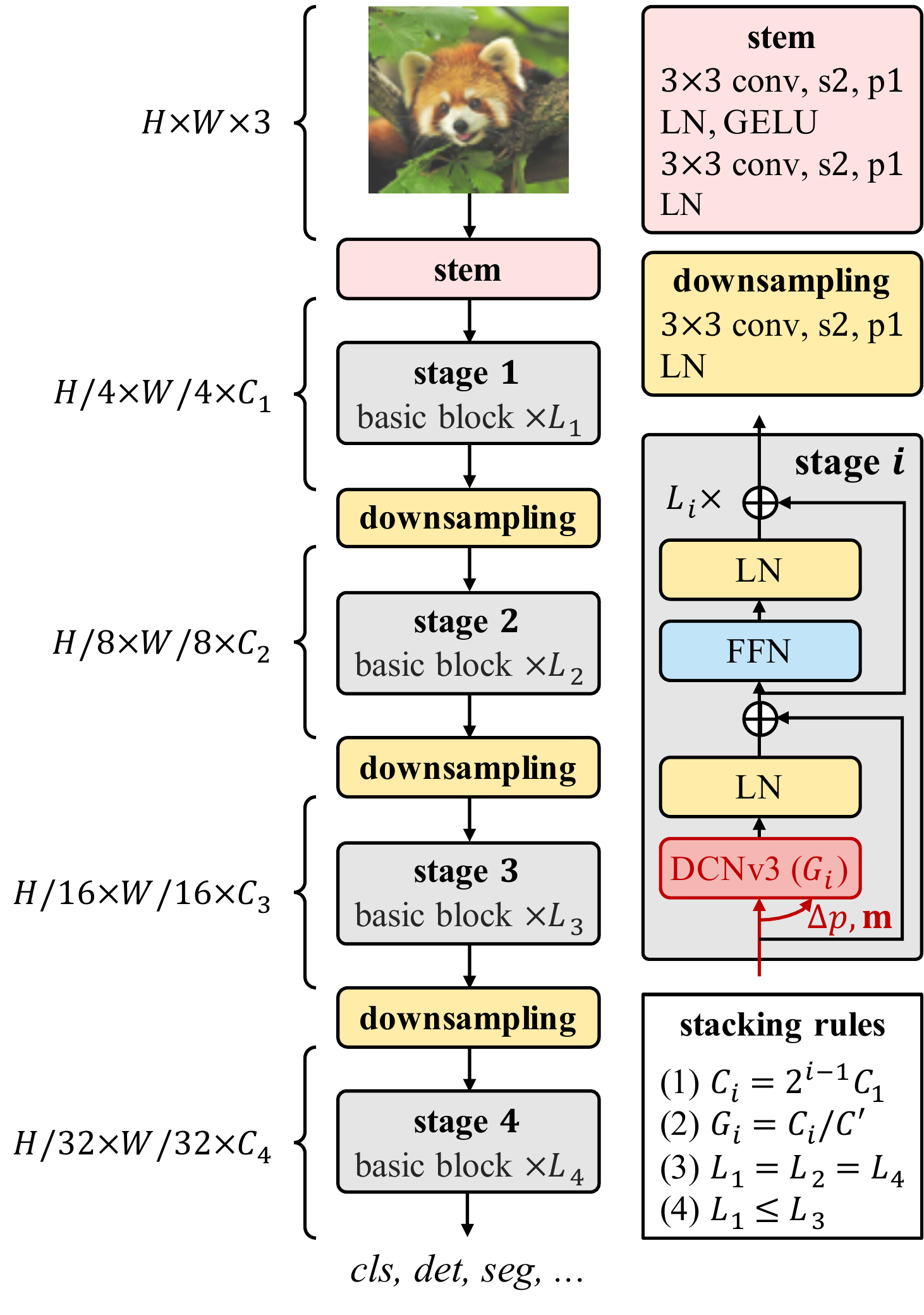}
    \caption{\textbf{Overall Architecture of InternImage}, where the core operator is DCNv3, and the basic block composes of layer normalization (LN)~\cite{ba2016layernorm} and feed-forward network (FFN)~\cite{vaswani2017attention} as transformers, the stem and downsampling layers follows conventional CNN's designs, where ``s2'' and ``p1'' mean stride 2 and padding 1, respectively. Constrained by the stacking rules, only 4 hyper-parameters $(C_1, C', L_1, L_3)$ can decide a model variant.}
    \label{fig:arch}
\end{figure}

(1) \emph{Sharing weights among convolutional neurons.} Similar to regular convolution, different convolutional neurons\footnote{A 3$\times$3 regular convolution has 9 linear projection neurons.} in original DCNv2 have independent linear projection weights,
and thus its parameter and memory complexity are linear with the total number of sampling points, which significantly limits the efficiency of the model, especially in large-scale models.
To remedy this problem, we borrow the idea from the separable convolution~\cite{chollet2017xception} and detach the original convolution weights $\mathbf{w}_k$ into depth-wise and point-wise parts,
where the depth-wise part is responsible by the original location-aware modulation scalar $\mathbf{m}_{k}$, and the point-wise part is the shared projection weights $\mathbf{w}$ among sampling points.

(2) \emph{Introducing multi-group mechanism}. The multi-group (head) design first appeared in group convolution~\cite{krizhevsky2017imagenet}, and
it is widely used in MHSA~\cite{vaswani2017attention} of transformers and works with adaptive spatial aggregation to effectively learn richer information from different representation subspaces at different locations.
Inspired by this, we split the spatial aggregation process into $G$ groups, each of which has individual sampling offsets $\Delta p_{gk}$ and modulation scale $\mathbf{m}_{gk}$,
and thus different groups on a single convolution layer can have different spatial aggregation patterns,
resulting in stronger features for downstream tasks.

(3) \emph{Normalizing modulation scalars along sampling points.} 
The modulation scalars in the original DCNv2 are element-wise normalized by the sigmoid function. 
Therefore, each modulation scalar is in the range [0, 1], and the sum of the modulation scalars of all sample points is not stable and varies from 0 to $K$.
This leads to unstable gradients in DCNv2 layers when training with large-scale parameters and data.
To alleviate the instability issues,
we change element-wise sigmoid normalization to softmax normalization along sample points.
In this way, the sum of the modulation scalars is constrained to 1, which makes the training process of models at different scales more stable.

Combining the aforementioned modifications, the extended DCNv2, marked as DCNv3, 
can be formulated as Eqn. \eqref{eqn:dcnv3}.
\begin{equation}
    \textbf{y}(p_0) = \sum^{G}_{g=1} \sum^{K}_{k=1} \mathbf{w}_g \mathbf{m}_{gk}\mathbf{x}_g(p_0 + p_k + \Delta p_{gk}),
    \label{eqn:dcnv3}
\end{equation}
where $G$ denotes the total number of aggregation groups.
For the $g$-th group,
$\mathbf{w}_g\!\in\!\mathbb{R}^{C\times C'}$ denotes the location-irrelevant projection weights of the group, where $C'\!=\!C/G$ represents the group dimension.
$\mathbf{m}_{gk}\!\in\!\mathbb{R}$ denotes the modulation scalar of the $k$-th sampling point in the $g$-th group, normalized by the softmax function along the dimension $K$.
$\mathbf{x}_g\!\in\!\mathbb{R}^{C'\times H \times W}$ represents the sliced input feature map.
$\Delta p_{gk}$ is the offset corresponding to the grid sampling location $p_k$ in the $g$-th group.

In general, DCNv3, as an extension of the DCN series, enjoys three merits as follows:
(1) This operator made up for the deficiencies of regular convolution in terms of long-range dependencies and adaptive spatial aggregation; 
(2) Compared with attention-based operators such as common MHSA and closely-related deformable attention \cite{zhu2020deformabledetr,xia2022vision}, this operator inherits the inductive bias of convolution, making our model more efficient with fewer training data and shorter training time;
(3) This operator is based on sparse sampling, which is more computational and memory efficient than previous methods such as MHSA~\cite{vaswani2017attention} and re-parameterizing large kernel~\cite{ding2022replknet}. In addition, due to the sparse sampling, DCNv3 only needs a 3$\times$3 kernel to learn long-range dependencies, which is easier to be optimized and avoids extra auxiliary techniques such as re-parameterizing~\cite{ding2022replknet} used in large kernels.

\subsection{\model\ Model}
Using DCNv3 as the core operator brings a new problem: \emph{how to build a model that can make effective use of the core operator?}
In this section, we first present the details of the basic block and other integral layers of our model, and then we construct a new CNN-based foundation model termed \model, by exploring a tailored stacking strategy for these basic blocks. Finally, we study scaling-up rules for the proposed model to obtain the gain from increasing parameters.

\textbf{Basic block.}
Unlike the widely used bottlenecks in traditional CNNs~\cite{he2016deep}, the design of our basic block is closer to ViTs, which is equipped with more advanced components including LN~\cite{ba2016layernorm}, feed-forward networks (FFN)~\cite{vaswani2017attention}, and GELU~\cite{hendrycks2016gelu}.
This design is proved to be efficient~\cite{wang2021pyramid,wang2021pvtv2,liu2021swin,liu2022convnet,ding2022replknet} in various vision tasks.
The details of our basic block are illustrated in Fig. \ref{fig:arch}, where the core operator is DCNv3,
and the sampling offsets and modulation scales are predicted by passing input feature $\mathbf{x}$ through a separable convolution (a 3$\times$3 depth-wise convolution followed by a linear projection).
For other components, we use the post-normalization setting~\cite{xiong2020layer} by default and follow the same design as that of the plain transformer~\cite{vaswani2017attention,dosovitskiy2020image}.

\textbf{Stem \& downsampling layers.}
To obtain hierarchical feature maps, we use 
convolutional stem and downsampling layers to resize the feature maps to different scales.
As shown in Fig. \ref{fig:arch}, the stem layer is placed before the first stage to reduce the input resolution by 4 times.
It consists of two convolutions, two LN layers, and one GELU layer,
where the kernel size of the two convolutions is 3, the stride is 2, the padding is 1,
and the output channel of the first convolution is half of the second one.
Similarly, the downsampling layer is made up of a 3$\times$3 convolution with a stride of 2 and a padding of 1, followed by one LN layer.
It sits between the two stages and is used to downsample the input feature map by 2 times.

\textbf{Stacking rules.}
To clarify the block-stacking process, we first list the integral hyperparameters of the \model\ as follows:\\
\indent $C_i$: the channel number of the $i$-th stage;\\
\indent $G_i$: the group number of the DCNv3 in the $i$-th stage;\\
\indent$L_i$: the number of basic blocks in the $i$-th stage.\\
Since our model has 4 stages, a variant is decided by 12 hyper-parameters,
whose search space is too large to exhaustively enumerate and find the best variant.
To reduce the search space, we summarize the design experiences of prior arts~\cite{he2016deep,liu2021swin,liu2022convnet} into 4 rules as shown in Fig.~\ref{fig:arch},
where the first rule makes the channel numbers of the last three stages determined by the channel number $C_1$ of the first stage,
and the second rule lets the group number correspond to the channel number of stages.
For the number of stacked blocks in different stages, we simplify the stacking pattern to ``AABA'', which means the block number of stage 1, 2, and 4 are the same, and are not greater than that of the stage 3 as illustrated in the last two rules.
With these rules, a \model\ variant can be defined by using only 4 hyper-parameters $(C_1, C', L_1, L_3)$.

Let us choose a model with 30 million parameters as the origin and discretize $C_1$ to $\{48, 64, 80\}$, $L_1$ to $\{1, 2, 3, 4, 5\}$, and $C'$ to $\{16, 32\}$.
In this way, the original huge search space is reduced to 30,
and we can find the best model from the 30 variants by training and evaluating them in ImageNet~\cite{deng2009imagenet}.
In practice, we use the best hyper-parameter setting $(64, 16, 4, 18)$ to define the origin model and scale it to different scales.

\begin{table}[t]
    \centering
    \renewcommand\arraystretch{1.0}
    \setlength{\tabcolsep}{2.8mm}
    \footnotesize
    \begin{tabular}{l|c|c|c|c}
    \renewcommand{\arraystretch}{0.1}
	model name & $C_1$ & $C'$  & $L_{1, 2, 3, 4}$ & \#params \\
	\hline
        \model-T (origin) &64 & 16 & 4, 4, 18, 4 & 30M \\
        \model-S &80 & 16 & 4, 4, 21, 4 & 50M \\
        \model-B &112 & 16 & 4, 4, 21, 4 & 97M \\
        \model-L &160 & 16 & 5, 5, 22, 5 & 223M \\
        \model-XL &192 & 16 & 5, 5, 24, 5 & 335M \\
        \model-H &320 & 32 & 6, 6, 32, 6 & 1.08B \\
    
\end{tabular}
    \caption{\textbf{Hyper-parameters for models of different scales}. \model-T is the origin model, and -S/B/L/XL/H are scaled up from -T. ``\#params'' denotes the number of parameters.}
    \label{tab:model_size}
\end{table}

\textbf{Scaling rules.} Based on the optimal origin model under the aforementioned constraints, we further explore the parameter scaling rules inspired by \cite{tan2019efficientnet}.
Specifically, we consider two scaling dimensions: depth $D$ (\ie, $3L_1\!+\!L_3$) and width $C_1$,
and scale the two dimensions using $\alpha$, $\beta$ and a composite factor $\phi$. The scaling rules can be written as:
$D'\!=\!\alpha^\phi D$ and $C'_1\!=\!\beta^\phi C_1$,
where $\alpha\!\geq\!1$, $\beta\!\geq\!1$, and $\alpha\beta^{1.99}\!\approx\!2$. Here, 1.99 is specific for InternImage and calculated by doubling the model width and keeping the depth constant.
We experimentally find out that the best scaling setting is $\alpha\!=\!1.09$ and $\beta\!=\!1.36$, and then we base on it to construct \model\ variants with different parameter scales, namely \model-T/S/B/L/XL, whose complexity is similar to those of ConvNeXt~\cite{liu2022convnet}. 
To further test the capability, we built a larger \model-H with 1 billion parameters, and to accommodate very large model widths, we also change the group dimension $C'$ to 32.
The configurations are summarized in Table~\ref{tab:model_size}.

\section{Experiment}
We analyze and compare \model\ with the leading CNNs and ViTs on representative vision tasks including image classification, object detection, instance and semantic segmentation.
Besides the experiments in the main paper, due to space constraints, more experimental setups and ablation studies are presented in the supplementary material.

\subsection{Image Classification}

\textbf{Settings.}
We evaluate the classification performance of \model\ on ImageNet~\cite{deng2009imagenet}. 
For fair comparisons, following common practices~\cite{touvron2021training,wang2021pyramid,liu2021swin,liu2022convnet}, \model-T/S/B are trained on ImageNet-1K ($\sim$1.3 million) for 300 epochs, 
and \model-L/XL are first trained on ImageNet-22K ($\sim$14.2 million) for 90 epochs and then fine-tuned on ImageNet-1K for 20 epochs.
To further explore the capability of our model and match the large-scale private data used in previous methods~\cite{dai2021coatnet,liu2021swinv2,yuan2021florence}, we adopt M3I Pre-training~\cite{su2022towards}, a unified pre-training approach available for both unlabeled and weakly-labeled data, to pre-train \model-H on a 427 million joint dataset of public Laion-400M~\cite{schuhmann2021laion}, YFCC-15M~\cite{thomee2016yfcc100m}, and CC12M~\cite{changpinyo2021conceptual} for 30 epochs,
and then we fine-tune the model on ImageNet-1K for 20 epochs.

\begin{table}[t!]
    \centering
    \renewcommand\arraystretch{0.89}
    \setlength{\tabcolsep}{0.9mm}
    \footnotesize
    \begin{tabular}{l|c|c|c|c|c}
    method & type & scale & \#params & \#FLOPs & acc (\%)  \\
    \hline
    DeiT-S~\cite{touvron2021training} & T &$224^2$  & 22M & 5G & 79.9 \\
    PVT-S~\cite{wang2021pyramid} & T &$224^2$  & 25M & 4G & 79.8 \\
    Swin-T~\cite{liu2021swin} & T &$224^2$ & 29M & 5G & 81.3\\
    CoAtNet-0~\cite{dai2021coatnet} & T &$224^2$ & 25M & 4G & 81.6 \\
    CSwin-T~\cite{dong2021cswin} & T &$224^2$  & 23M & 4G & 82.7\\
    PVTv2-B2~\cite{wang2021pvtv2} & T & $224^2$  & 25M & 4G & 82.0 \\
    DeiT III-S~\cite{touvron2022deit3} & T &$224^2$  & 22M & 5G & 81.4\\
    SwinV2-T/8~\cite{liu2021swinv2} & T &$256^2$ & 28M & 6G & 81.8\\
    Focal-T~\cite{yang2021focal} & T &$224^2$ & 29M & 5G & 82.2\\
    ConvNeXt-T~\cite{liu2022convnet} & C & $224^2$ & 29M & 5G & 82.1\\
    ConvNeXt-T-dcls~\cite{hassani2021dilated} & C & $224^2$ & 29M & 5G & 82.5\\
    SLaK-T~\cite{liu2022slak} & C & $224^2$  & 30M & 5G & 82.5\\
    HorNet-T~\cite{rao2022hornet} & C & $224^2$ & 23M & 4G & 83.0 \\
    \rowcolor{gray!20}
    \model-T (ours) & C &$224^2$  & 30M & 5G & 83.5 \\
    \hline
    PVT-L~\cite{wang2021pyramid} & T &$224^2$  & 61M & 10G & 81.7 \\
    Swin-S~\cite{liu2021swin} & T &$224^2$  & 50M & 9G & 83.0\\
    CoAtNet-1~\cite{dai2021coatnet} & T &$224^2$ & 42M & 8G & 83.3 \\
    PVTv2-B4~\cite{wang2021pvtv2} & T & $224^2$  & 63M & 10G & 83.6 \\
    SwinV2-S/8~\cite{liu2021swinv2} & T &$256^2$ & 50M & 12G & 83.7\\
    ConvNeXt-S~\cite{liu2022convnet} & C &$224^2$  & 50M & 9G & 83.1\\
    ConvNeXt-S-dcls~\cite{hassani2021dilated} & C &$224^2$  & 50M & 10G & 83.7\\
    SLaK-S~\cite{liu2022slak} & C &$224^2$  & 55M & 10G & 83.8 \\
    HorNet-S~\cite{rao2022hornet} & C & $224^2$ & 50M & 9G & 84.0 \\
    \rowcolor{gray!20}
    \model-S (ours) & C &$224^2$  & 50M & 8G & 84.2 \\
    \hline
    DeiT-B~\cite{touvron2021training} & T &$224^2$ & 87M & 18G & 83.1 \\
    Swin-B~\cite{liu2021swin} & T &$224^2$  & 88M & 15G & 83.5\\
    CoAtNet-2~\cite{dai2021coatnet} & T &$224^2$ & 75M & 16G & 84.1 \\
    PVTv2-B5~\cite{wang2021pvtv2} & T & $224^2$  & 82M & 12G & 83.8 \\
    DeiT III-B~\cite{touvron2022deit3} & T &$224^2$ & 87M & 18G & 83.8 \\
    SwinV2-B/8~\cite{liu2021swinv2} & T &$256^2$  & 88M & 20G & 84.2\\
    RepLKNet-31B~\cite{ding2022replknet} & C &$224^2$  & 79M & 15G & 83.5\\
    ConvNeXt-B~\cite{liu2022convnet} & C &$224^2$  & 88M & 15G & 83.8\\
    ConvNeXt-B-dcls~\cite{hassani2021dilated} & C &$224^2$  & 89M & 17G & 84.1\\
    SLaK-B~\cite{liu2022slak} & C & $224^2$  & 95M & 17G & 84.0 \\
    HorNet-B~\cite{rao2022hornet} & C & $224^2$ & 88M & 16G & 84.3 \\
    \rowcolor{gray!20}
    \model-B (ours) & C &$224^2$ & 97M & 16G & 84.9 \\
    \hline
    Swin-L$^\ddagger$~\cite{liu2021swin} & T &$384^2$ & 197M & 104G & 87.3\\
    CoAtNet-3$^\ddagger$~\cite{dai2021coatnet} & T &$384^2$ & 168M & 107G & 87.6 \\
    CoAtNet-4$^\ddagger$~\cite{dai2021coatnet} & T &$384^2$ & 275M & 190G & 87.9 \\
    DeiT III-L$^\ddagger$~\cite{touvron2022deit3} & T &$384^2$ & 304M & 191G & 87.7 \\
    SwinV2-L/24$^\ddagger$~\cite{liu2021swinv2} & T &$384^2$ & 197M & 115G & 87.6\\
    RepLKNet-31L$^\ddagger$~\cite{ding2022replknet} & C &$384^2$ & 172M & 96G & 86.6\\
    HorNet-L$^\ddagger$~\cite{rao2022hornet} & C & $384^2$ & 202M & 102G & 87.7 \\
    ConvNeXt-L$^\ddagger$~\cite{liu2022convnet} & C &$384^2$ & 198M & 101G & 87.5\\
    ConvNeXt-XL$^\ddagger$~\cite{liu2022convnet} & C &$384^2$ & 350M & 179G & 87.8\\
    \rowcolor{gray!10}
    \model-L$^\ddagger$ (ours) & C &$384^2$ & 223M & 108G & 87.7 \\
    \rowcolor{gray!20}
    \model-XL$^\ddagger$ (ours) & C &$384^2$ & 335M & 163G & 88.0 \\
    \hline
    ViT-G/14$^\#$~\cite{zhai2022scaling} & T & $518^2$  & 1.84B & 5160G & 90.5\\
    CoAtNet-6$^\#$~\cite{dai2021coatnet} & T & $512^2$ & 1.47B & 1521G & 90.5\\
    CoAtNet-7$^\#$~\cite{dai2021coatnet} & T & $512^2$ & 2.44B & 2586G & 90.9\\
    Florence-CoSwin-H$^\#$~\cite{yuan2021florence} & T & $-$ & 893M & $-$ & 90.0 \\
    SwinV2-G$^\#$~\cite{liu2021swinv2} & T & $640^2$ & 3.00B & $-$ & 90.2 \\
    RepLKNet-XL$^\#$~\cite{ding2022replknet} & C & $384^2$ & 335M & 129G & 87.8\\
    BiT-L-ResNet152x4$^\#$~\cite{kolesnikov2020big} & C & $480^2$ & 928M & $-$ & 87.5 \\
    \rowcolor{gray!10}
    \model-H$^\#$ (ours) & C & $224^2$ & 1.08B & 188G & 88.9  \\
    \rowcolor{gray!20}
    \model-H$^\#$ (ours) & C & $640^2$ & 1.08B & 1478G & 89.6  \\
        
\end{tabular}
    \vspace{1pt}
    \caption{\textbf{Image classification performance on the ImageNet validation set}. ``type'' refers to model type, where ``T'' and ``C'' denote transformer and CNN, respectively. ``scale'' is the input scale. ``acc'' is the top-1 accuracy.
    ``$^\ddagger$" indicates the model is pre-trained on ImageNet-22K~\cite{deng2009imagenet}. ``$^\#$" indicates pretraining on extra large-scale private dataset such as JFT-300M~\cite{xie2020self}, FLD-900M~\cite{yuan2021florence}, or the joint public dataset in this work.}
    \label{tab:cls_imagenet}
\end{table}

\textbf{Results.} Table~\ref{tab:cls_imagenet} shows the classification results of models with different scales.
With similar parameters and computational costs, our models are comparable or even superior to the state-of-the-art transformer-based and CNN-based models. 
For example, \model-T achieves 83.5\% top-1 accuracy, outperforming ConvNext-T \cite{liu2022convnet} with a clear margin of 1.4 points.
\model-S/B keeps the leading position and \model-B surpasses the hybrid-ViT CoAtNet-2 \cite{dai2021coatnet} by 0.8 points.
When pre-trained on ImageNet-22K and the large-scale joint dataset, the top-1 accuracy of \model-XL and -H are boosted to 88.0\% and 89.6\%, respectively,
which is better than previous CNNs~\cite{kolesnikov2020big,ding2022replknet} also trained with large-scale data, and closes the gap with the state-of-the-art large-scale ViTs to about 1 point.
This gap may be caused by the discrepancy between large-scale inaccessible private data and the aforementioned joint public data.
These results show that our \model\ not only has good performance on the common parameter scale and the public training data, but also can effectively extend to large-scale parameters and data.

\def\detmodel{Intern}

\begin{table*}[t]\small
	\centering
	\footnotesize
    \setlength\tabcolsep{1.55mm}{
    \begin{tabular}{l|cc|cccccc|cccccc}
        \multirow{2}{*}{method} & \multirow{2}{*}{\#params} & \multirow{2}{*}{\#FLOPs} &
        \multicolumn{6}{c|}{Mask R-CNN 1$\times$ schedule} & \multicolumn{6}{c}{Mask R-CNN 3$\times$+MS schedule}\\
        ~ &  &
        & $\rm AP^b$ & $\rm AP^b_{50}$ & $\rm AP^b_{75}$ & $\rm AP^m$ & $\rm AP^m_{50}$ & $\rm AP^m_{75}$ 
        & $\rm AP^b$ & $\rm AP^b_{50}$ & $\rm AP^b_{75}$ & $\rm AP^m$ & $\rm AP^m_{50}$ & $\rm AP^m_{75}$   \\
	    \hline
        Swin-T~\cite{liu2021swin} & 48M & 267G
        & 42.7 & 65.2 & 46.8 & 39.3 & 62.2 & 42.2 & 46.0 & 68.1 & 50.3 & 41.6 & 65.1 & 44.9  \\
        ConvNeXt-T~\cite{liu2022convnet} & 48M & 262G
        & 44.2 & 66.6 & 48.3 & 40.1 & 63.3 & 42.8 & 46.2 & 67.9 & 50.8 & 41.7 & 65.0 & 44.9  \\
        PVTv2-B2~\cite{wang2021pvtv2} & 45M & 309G & 45.3 & 67.1 & 49.6 & 41.2 & 64.2 & 44.4 & 47.8 & 69.7 & 52.6 & 43.1 & 66.8 & 46.7  \\
        ViT-Adapter-S~\cite{chen2022vitadapter} & 48M & 403G & 44.7 & 65.8 & 48.3 & 39.9 & 62.5 & 42.8 & 48.2 & 69.7 & 52.5 & 42.8 & 66.4 & 45.9 \\
        \rowcolor{gray!20}
        \model-T (ours)  & 49M & 270G
        & 47.2 & 69.0 & 52.1 & 42.5 & 66.1 & 45.8 & 49.1 & 70.4 & 54.1 & 43.7 & 67.3 & 47.3  \\
        \hline
        Swin-S~\cite{liu2021swin} & 69M & 354G
        & 44.8 & 66.6 & 48.9 & 40.9 & 63.4 & 44.2 & 48.2 & 69.8 & 52.8 & 43.2 & 67.0 & 46.1 \\
        ConvNeXt-S~\cite{liu2022convnet} & 70M & 348G
        & 45.4 & 67.9 & 50.0 & 41.8 & 65.2 & 45.1 & 47.9 & 70.0 & 52.7 & 42.9 & 66.9 & 46.2 \\
        PVTv2-B3~\cite{wang2021pvtv2} & 65M & 397G & 47.0 & 68.1 & 51.7 & 42.5 & 65.7 & 45.7 & 48.4 & 69.8 & 53.3 & 43.2 & 66.9 & 46.7 \\
        \rowcolor{gray!20}
        \model-S (ours)  & 69M & 340G
        & 47.8 & 69.8 & 52.8 & 43.3 & 67.1 & 46.7 & 49.7 & 71.1 & 54.5 & 44.5 & 68.5 & 47.8 \\
        \hline
        Swin-B~\cite{liu2021swin} & 107M & 496G
        & 46.9 & $-$ & $-$ & 42.3 & $-$ & $-$ & 48.6 & 70.0 & 53.4 & 43.3 & 67.1 & 46.7  \\
        ConvNeXt-B~\cite{liu2022convnet} & 108M & 486G & 47.0 & 69.4 & 51.7 & 42.7 & 66.3 & 46.0 & 48.5 & 70.1 & 53.3 & 43.5 & 67.1 & 46.7 \\
        PVTv2-B5~\cite{wang2021pvtv2} & 102M & 557G
        & 47.4 & 68.6 & 51.9 & 42.5 & 65.7 & 46.0 & 48.4 & 69.2 & 52.9 & 42.9 & 66.6 & 46.2  \\
        ViT-Adapter-B~\cite{chen2022vitadapter}  & 120M & 832G & 47.0 & 68.2 & 51.4 & 41.8 & 65.1 & 44.9 & 49.6 & 70.6 & 54.0 & 43.6 & 67.7 & 46.9  \\
        \rowcolor{gray!20}
        \model-B (ours) & 115M & 501G
        & 48.8 & 70.9 & 54.0 & 44.0 & 67.8 & 47.4 & 50.3 & 71.4 & 55.3 & 44.8 & 68.7 & 48.0 \\
        \multicolumn{15}{c}{ } \\
        method & \#param & \#FLOPs & \multicolumn{6}{c|}{Cascade Mask R-CNN 1$\times$ schedule} & \multicolumn{6}{c}{Cascade Mask R-CNN 3$\times$+MS schedule}\\
        \hline
        Swin-L$^\ddagger$~\cite{liu2021swin} & 253M & 1382G & 51.8 & 71.0 & 56.2 & 44.9 & 68.4 & 48.9 & 53.9 & 72.4 & 58.8 & 46.7 & 70.1 & 50.8 \\
        ConvNeXt-L$^\ddagger$~\cite{liu2022convnet} & 255M & 1354G & 53.5 & 72.8 & 58.3 & 46.4 & 70.2 & 50.2 & 54.8 & 73.8 & 59.8 & 47.6 & 71.3 & 51.7  \\
        RepLKNet-31L$^\ddagger$~\cite{ding2022replknet} & 229M & 1321G & $-$ & $-$  & $-$ & $-$ & $-$ & $-$ & 53.9 & 72.5 & 58.6 & 46.5 & 70.0 & 50.6 \\
        HorNet-L$^\ddagger$~\cite{rao2022hornet} & 259M & 1358G & $-$ & $-$  & $-$ & $-$ & $-$ & $-$ & 56.0 & $-$ & $-$ & 48.6 & $-$ & $-$ \\
        \rowcolor{gray!20}
        \model-L$^\ddagger$ (ours) & 277M & 1399G & 54.9 & 74.0 & 59.8 & 47.7 & 71.4 & 52.1 & 56.1 & 74.8 & 60.7 & 48.5 & 72.4 & 53.0 \\
        \hline
        ConvNeXt-XL$^\ddagger$~\cite{liu2022convnet} & 407M & 1898G & 53.6 & 72.9 & 58.5 & 46.5 & 70.3 & 50.5 & 55.2 & 74.2 & 59.9 & 47.7 & 71.6 & 52.2\\
        \rowcolor{gray!20}
        \model-XL$^\ddagger$ (ours) & 387M & 1782G & 55.3 & 74.4 & 60.1 & 48.1 & 71.9 & 52.4 & 56.2 & 75.0 & 61.2 & 48.8 & 72.5 & 53.4 \\  
    \end{tabular}}
    \caption{
    \textbf{Object detection and instance segmentation performance on COCO \texttt{val2017}.}
    The FLOPs are measured with 1280$\times$800 inputs.
    AP$^\text{b}$ and AP$^\text{m}$ represent box AP and mask AP, respectively.
    ``MS" means multi-scale training. }
    \label{tab:results_detection_mask}
    \label{tab:mask}
\end{table*}

\subsection{Object Detection}
\label{object_detection_exp}

\textbf{Settings.}
We verify the detection performance of our \model\ on the COCO benchmark~\cite{lin2014microsoft}, on top of two representative object detection frameworks: Mask R-CNN~\cite{he2017mask}, and Cascade Mask R-CNN~\cite{cai2019cascade}.
We follow common practices~\cite{liu2021swin,wang2021pvtv2} to initialize the backbone with pre-trained classification weights, and train models use a 1$\times$ (12 epochs) or 3$\times$ (36 epochs) schedule by default.

\textbf{Results.}
As shown in Table~\ref{tab:results_detection_mask}, when using Mask R-CNN for object detection, 
we find that under a comparable number of parameters, our models significantly surpass their counterparts. 
For example, with the $1\times$ training schedule, the box AP (AP$^{\rm b}$) of \model-T is 4.5 points better than Swin-T~\cite{liu2021swin} (47.2 \vs 42.7),
and 3.0 points higher than ConvNeXt-T~\cite{liu2022convnet} (47.2 \vs 44.2).
With the 3$\times$ multi-scale training schedule, more parameters, and more advanced Cascade Mask R-CNN~\cite{cai2019cascade},
\model-XL achieves AP$^{\rm b}$ of 56.2, surpassing ConvNeXt-XL by 1.0 points (56.2 \vs 55.2). 
Similar results are also seen in instance segmentation experiments.
With the 1$\times$ training schedule, \model-T yields 42.5 mask AP (\ie, AP$^{\rm m}$), 
which outperforms Swin-T and ConvNeXt-T by 3.2 points (42.5 \vs 39.3) and 2.4 points (42.5 \vs 40.1), respectively.
The best AP$^{\rm m}$ 48.8 is obtained by \model-XL with Cascade Mask R-CNN, which is at least 1.1 points higher than its counterparts.

\def\detmodel{Intern}

\begin{table}[t]\small
	\centering
	\footnotesize
    \setlength\tabcolsep{0.1mm}{
    \begin{tabular}{l|l|c|cc}
        \multirow{2}{*}{method} & \multirow{2}{*}{detector} & \multirow{2}{*}{\#params} & \multicolumn{2}{c}{$\rm AP^\text{b}$} \\
        &  & & val2017 & test-dev \\
	    \hline
        Swin-L~\cite{liu2021swin} & DyHead~\cite{dai2021dynamic} & 213M & 56.2 & 58.4 \\
	            
    	Swin-L$^\ddagger$~\cite{liu2021swin} 
    	& HTC++~\cite{liu2021swin} & 284M & 58.0 & 58.7 \\

        Swin-L$^\ddagger$~\cite{liu2021swin} 
        & Soft-Teacher~\cite{xu2021end} & 284M & 60.7 & 61.3 \\
        
        Florence-CoSwin-H$^\#$~\cite{yuan2021florence} 
        & DyHead~\cite{dai2021dynamic} & 
        637M
        & 62.0 & 62.4 \\
        
        ViT-L$^\ddagger$~\cite{dosovitskiy2020image} 
        & ViT-Adapter~\cite{chen2022vitadapter} & 401M & 62.6 & 62.6 \\

        Swin-L$^\ddagger$~\cite{liu2021swin} 
        & DINO~\cite{zhang2022dino} & 218M & 63.2 & 63.3 \\

        FocalNet-H$^\ddagger$~\cite{yang2022focal} 
        & DINO~\cite{zhang2022dino} & 746M & 64.2 & 64.3 \\
        
        ViT-Huge~\cite{chen2022group} 
        & Group-DETRv2~\cite{chen2022group} & 629M & $-$ & 64.5 \\

        SwinV2-G$^\#$\cite{liu2021swinv2} 
        & HTC++~\cite{liu2021swin} & 3.00B & 62.5 & 63.1 \\
        
        BEiT-3$^\#$~\cite{wang2022beit3} 
        & ViTDet~\cite{li2022exploring} & 1.90B &$-$ & 63.7 \\
        
        FD-SwinV2-G$^\#$~\cite{wei2022fdswin}
        & HTC++~\cite{liu2021swin} & 3.00B & $-$ & 64.2 \\
        \hline
        \rowcolor{gray!10}
        \model-XL$^\ddagger$ (ours) 
        & DINO~\cite{zhang2022dino} & 602M & 64.2 & 64.3 \\ 
        \rowcolor{gray!20}
        \model-H$^\#$ (ours) 
        & DINO~\cite{zhang2022dino} & 2.18B & 65.0 & 65.4\\
    \end{tabular}
    }
    \vspace{-3pt}
    \caption{\textbf{Comparison of the state-of-the-art detectors on COCO \texttt{val2017} and test-dev.}
    \vspace{-10pt}
    }
    \label{tab:sota}
\end{table}

To further push the performance bound of object detection, we follow the advanced setting used in leading methods~\cite{wang2022beit3,liu2021swinv2,zhang2022dino,wei2022fdswin,liang2022cbnet} to initialize the backbone with the weights pre-trained on ImageNet-22K or the large-scale joint dataset, and double its parameters via the composite techniques~\cite{liang2022cbnet} (see the model with 2 billion parameters in Fig. \ref{fig:bb_cmp}).
Then, we fine-tune it along with the DINO~\cite{zhang2022dino} detector on the Objects365~\cite{shao2019objects365} and COCO datasets one after another for 26 epochs and 12 epochs, respectively.
As shown in Table~\ref{tab:sota}, our method achieves the best results of 65.0 AP$^\text{b}$ and 65.4 AP$^\text{b}$ on COCO val2017 and test-dev.
Compared to previous state-of-the-art models, we surpass FD-SwinV2-G~\cite{wei2022fdswin} by 1.2 points (65.4 \vs 64.2), with 27\% fewer parameters and without complicated distillation processes,
which shows the effectiveness of our models on the detection task.

\begin{table}[t]
    \centering
    \setlength{\tabcolsep}{1.3mm}
    \footnotesize
    \begin{tabular}{l|c|c|c|cc}
    	\multirow{2}{*}{method} & crop & \multirow{2}{*}{\#params} & \multirow{2}{*}{\#FLOPs} & mIoU & mIoU\\
    	& size & & & (SS) & (MS)   \\
    	\whline
    	Swin-T~\cite{liu2021swin} & 512$^2$ & 60M  & 945G & 44.5 & 45.8 \\
    	ConvNeXt-T~\cite{liu2022convnet} & 512$^2$ & 60M & 939G & 46.0 & 46.7 \\
    	SLaK-T~\cite{liu2022slak} & 512$^2$ & 65M & 936G & 47.6 & $-$ \\
    	\rowcolor{gray!20}
    	\model-T (ours) & 512$^2$ & 59M & 944G & 47.9 & 48.1  \\
    	\hline
        Swin-S~\cite{liu2021swin} & 512$^2$ & 81M &  1038G &  47.6 &  49.5 \\
        ConvNeXt-S~\cite{liu2022convnet}  & 512$^2$ &82M & 1027G  & 48.7 & 49.6  \\
        SLaK-S~\cite{liu2022slak} & 512$^2$ &91M & 1028G & 49.4 & $-$ \\
        \rowcolor{gray!20}
        \model-S (ours) & 512$^2$ & 80M & 1017G & 50.1 & 50.9 \\
        \hline
        Swin-B~\cite{liu2021swin} 
        & 512$^2$ & 121M & 1188G & 48.1 & 49.7  \\
        ConvNeXt-B~\cite{liu2022convnet}  
        & 512$^2$ & 122M & 1170G & 49.1 & 49.9  \\
        RepLKNet-31B~\cite{ding2022replknet} 
        & 512$^2$ & 112M & 1170G & 49.9 & 50.6  \\
        SLaK-B~\cite{liu2022slak} & 512$^2$ & 135M & 1172G & 50.2 & $-$    \\
        \rowcolor{gray!20}
        \model-B (ours) 
        & 512$^2$ & 128M & 1185G & 50.8 & 51.3 \\
        \hline
        Swin-L$^\ddagger$~\cite{liu2021swin} 
        & 640$^2$ & 234M & 2468G & 52.1 & 53.5 \\
        RepLKNet-31L$^\ddagger$~\cite{ding2022replknet} 
        & 640$^2$ & 207M & 2404G & 52.4 & 52.7 \\
        ConvNeXt-L$^\ddagger$~\cite{liu2022convnet} 
        & 640$^2$ & 235M & 2458G & 53.2 & 53.7 \\
        ConvNeXt-XL$^\ddagger$~\cite{liu2022convnet} 
        & 640$^2$ & 391M & 3335G & 53.6 & 54.0 \\
        \rowcolor{gray!10}
        \model-L$^\ddagger$ (ours) 
        & 640$^2$ & 256M  & 2526G  & 53.9 & 54.1 \\
        \rowcolor{gray!20}
        \model-XL$^\ddagger$ (ours)
        & 640$^2$ & 368M & 3142G & 55.0 & 55.3 \\
        \hline
        SwinV2-G$^\#$~\cite{liu2021swinv2}
        & 896$^2$ & 3.00B & $-$ & $-$ & 59.9 \\
        \rowcolor{gray!10}
        \model-H$^\#$ (ours)
        & 896$^2$ & 1.12B & 3566G & 59.9 & 60.3 \\
        \hline
        BEiT-3$^\#$~\cite{wang2022beit3} & 896$^2$ & 1.90B & $-$ & $-$ & 62.8 \\ 
        FD-SwinV2-G$^\#$~\cite{wei2022fdswin}
        & 896$^2$ & 3.00B & $-$ & $-$ & 61.4 \\
        \rowcolor{gray!20}
        \model-H$^\#$ (ours) + &  &  &  &  &  \\
        \rowcolor{gray!20}
        Mask2Former~\cite{cheng2021masked}
        & \multirow{-2}{*}{896$^2$}& \multirow{-2}{*}{1.31B}& \multirow{-2}{*}{4635G}& \multirow{-2}{*}{62.5}& \multirow{-2}{*}{62.9}\\
    \end{tabular}
    \caption{\textbf{Semantic segmentation performance on the ADE20K validation set.}
    The FLOPs are measured with 512$\times$2048, 640$\times$2560, or 896$\times$896 inputs according to the crop size.
    ``SS'' and ``MS" means single-scale and multi-scale testing, respectively. 
    }
    \label{tab:seg}
\end{table}

\subsection{Semantic Segmentation}

\textbf{Settings.}
To evaluate the semantic segmentation performance of \model,
we initialize the backbone with pre-trained classification weights and train our models with UperNet~\cite{xiao2018unified} 
on ADE20K~\cite{zhou2017scene} for 160k iterations and compare fairly with previous CNN-based and transformer-based backbones.
To further reach top performance,
we arm  
InternImage-H with more advanced Mask2Former~\cite{cheng2021masked}, and adopt the same training settings in \cite{chen2022vitadapter,wang2022beit3}.

\textbf{Results.}
As shown in Table~\ref{tab:seg}, when using UperNet \cite{xiao2018unified} for semantic segmentation, our \model~consistently outperforms prior arts \cite{liu2021swin, liu2022convnet, liu2022slak, ding2022replknet}.
For example, with almost the same parameter numbers and FLOPs, our \model-B reports 50.8 mIoU on the ADE20K val, which is outstanding from the strong counterparts such as ConvNeXt-B (50.8 \emph{vs.}~49.1) and RepLKNet-31B (50.8 \emph{vs.}~49.9).
Furthermore, our \model-H yields 60.3 MS mIoU, which is better than SwinV2-G \cite{liu2021swinv2}, while the parameter number is much smaller (1.12B \emph{vs.}~3.00B).

It is worth noting that, when using Mask2Former~\cite{cheng2021masked} and multi-scale testing, our \model-H achieves the best mIoU of 62.9, higher than the current best BEiT-3~\cite{wang2022beit3} on the ADE20K benchmark. 
These results demonstrate that the CNN-based foundation model can also enjoy the dividends of massive data and challenge the leading position of transformer-based models.

\subsection{Ablation Study}

\begin{figure}[t]
    \centering
    \includegraphics[width=1.0\columnwidth]{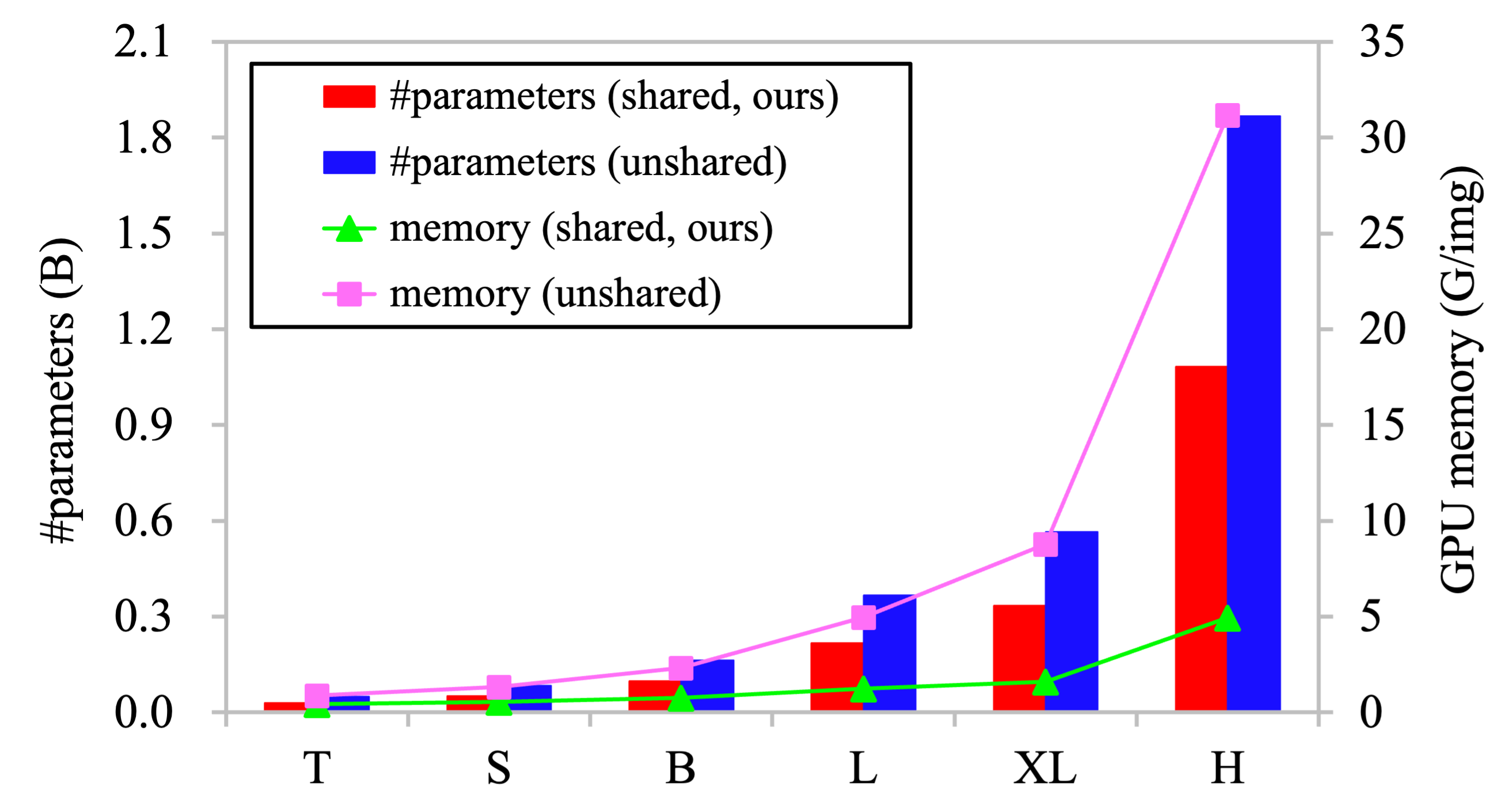}
    \caption{\textbf{Model parameters and GPU memory usage of shared weights \textit{v.s} unshared weights among convolution neurons.} 
    The left vertical axis indicates the model parameters and the right one indicates the GPU memory usage per image when the batch size is 32 and the input image resolution is $224\times224$.
    }
    \label{fig:spw}
\end{figure}

\begin{figure}[t]
    \centering
    \includegraphics[width=0.98\columnwidth]{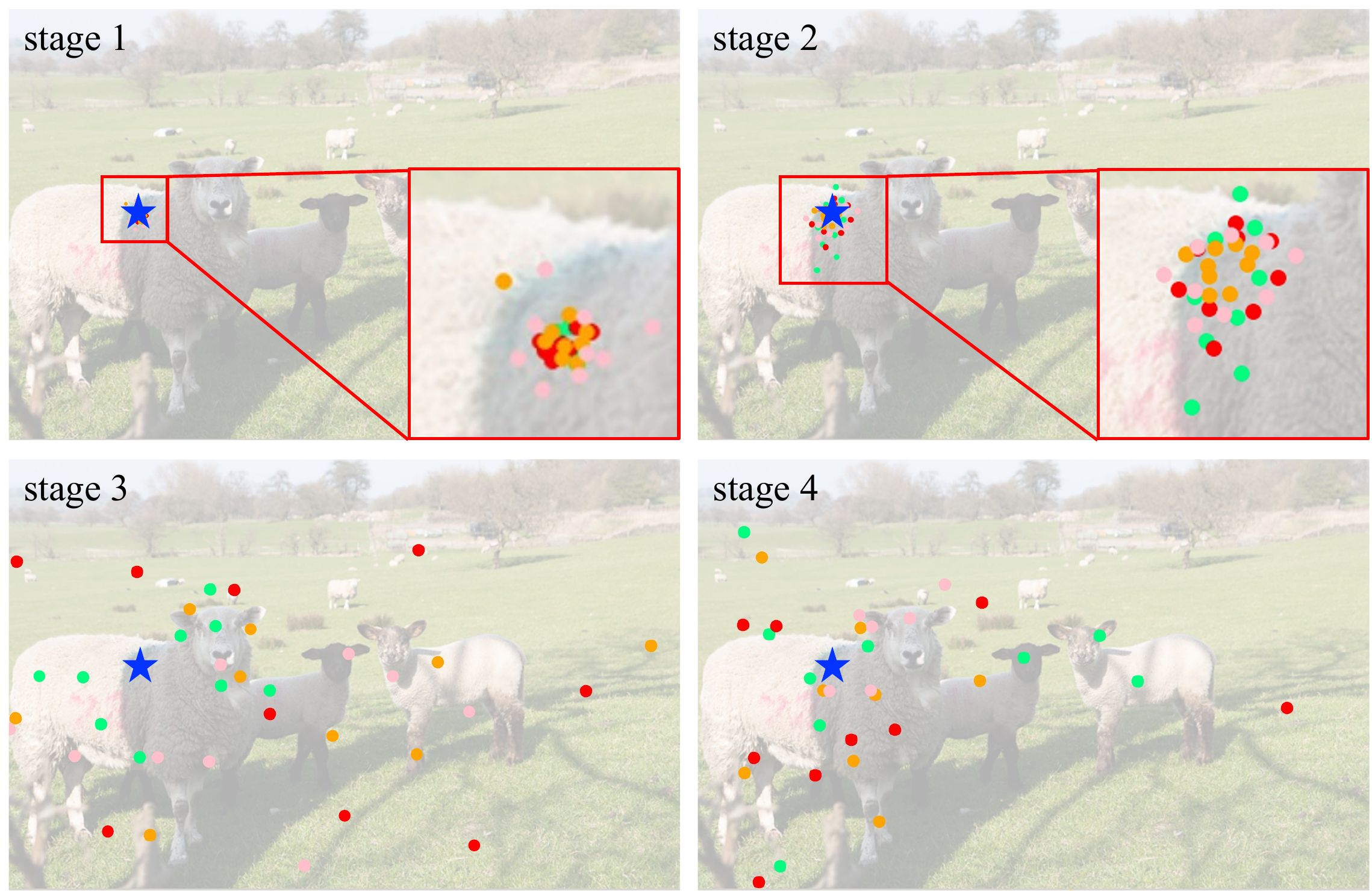}
    \caption{\textbf{Visualization of sampling locations for different groups at different stages}. The blue star indicates the query point (on the left sheep), 
    and the dots with different colors indicate the sampling locations of different
groups.
    }
    \label{fig:group}
\end{figure}

\textbf{Sharing weights among convolution neurons matters.}
Large-scale models are sensitive to 
parameters and memory cost of the core operator, due to hardware limitations.
To address this problem, we share weights among convolution neurons of DCNv3.
As shown in Fig. \ref{fig:spw}, we compare the parameters and memory cost of the models based on DCNv3 with shared or unshared weights.
We see that the parameters and memory cost of models with unshared weights are much higher than the shared one,
especially for the -H scale, the ratio of saved parameters and GPU memory is 42.0\% and 84.2\%, respectively.
As shown in Table \ref{tab:ablation_all}, we also examine that the two models at -T scale have similar top-1 accuracy on ImageNet (83.5 \vs 83.6) and AP$^\text{b}$ on COCO (47.2 \vs 47.4), even the model without shared weights has 66.1\% more parameters.

\textbf{Multi-group spatial aggregation brings stronger features.} We introduce aggregation groups to allow our model to learn information from different representation subspaces like transformers~\cite{dosovitskiy2020image}.
As shown in Fig. \ref{fig:group}, for the same query pixel, the offsets from different groups are concentrated in different regions, resulting in hierarchical semantic features.
We also compare the performance of the model with and without multiple groups. As reported in Table \ref{tab:ablation_all}, the model significantly drops 1.2 points on ImageNet and 3.4 points on COCO val2017.
In addition,
we also see that in the first two stages, the learned effective receptive field (ERF) is relatively small,
and as the model goes deeper (\ie, stages 3 and 4), the ERF increases to be global.
This phenomenon is different from ViTs~\cite{dosovitskiy2020image,wang2021pyramid,xie2021segformer} whose ERF is usually global.

\begin{table}[t]
    \centering
    \renewcommand{\arraystretch}{0.98}
    \setlength{\tabcolsep}{1.8mm}
    \footnotesize
    \begin{tabular}{ccc|c|cc}
        shared $\mathbf{w}$ & multi-group & softmax & top-1 acc & $\rm AP^b$ & $\rm AP^m$\\
        \whline
        \no & \yes & \yes & 83.6 & 47.4 & 42.6\\
        \yes & \no & \yes & 82.3 & 43.8 & 40.0\\
        \yes  &  \yes & \no & 65.7 & 38.7 & 35.6\\
        \yes  &  \yes & \yes & 83.5 & 47.2 & 42.5 \\
        \hline
    \end{tabular}
    \caption{\textbf{Ablation comparison of the three modifications in DCNv3. }
    These experiments are based on \model-T for classification and Mask R-CNN 1$\times$ schedule for detection.
    }
    \label{tab:ablation_all}
\end{table}

\section{Conclusion \& Limitations}
We introduce \model, a new large-scale CNN-based foundation model that can provide strong representations for versatile vision tasks, such as image classification, object detection, and semantic segmentation.
We tune the flexible DCNv2 operator to satisfy the requirement of foundation models, and develop a series of blocks, stacking and scaling rules centered on the core operator.
Extensive experiments on object detection and semantic segmentation benchmarks verify that our \model\ can obtain comparable or better performance than well-designed large-scale vision transformers trained with massive data,
showing that CNN is also a considerable choice for large-scale vision foundation model research.
Nonetheless, latency remains an issue for DCN-based operators adapting to downstream tasks with high-speed requirements.
Also, large-scale CNNs are still in their early stages of development, and we hope \model\ can serve as a good starting point.

\clearpage

\noindent\textbf{\Large Appendix}
\thispagestyle{empty}

\appendix

\section{Detailed Training Settings}

In this section, we present the detailed training recipes for image classification, object detection, and semantic segmentation.

\subsection{Settings for Backbone-Level Comparison}

\textbf{ImageNet image classification.} The training details of image classification on ImageNet~\cite{deng2009imagenet} are shown in Table \ref{tab:supp_cls_setting}, which are similar to common practices~\cite{liu2021swin,touvron2022deit3,touvron2021training,liu2022convnet} and with some tweaks. 
To further explore the capability of our model and match the large-scale private data used in previous methods~\cite{dai2021coatnet,liu2021swinv2,yuan2021florence}, we adopt M3I Pre-training~\cite{su2022towards}, a unified pre-training approach available for both unlabeled and weakly-labeled data, to pre-train \model-H on a 427 million joint dataset of public Laion-400M~\cite{schuhmann2021laion}, YFCC-15M~\cite{thomee2016yfcc100m}, and CC12M~\cite{changpinyo2021conceptual} for 30 epochs,
and then we fine-tune the model on ImageNet-1K for 20 epochs.
For the more detailed pre-training settings of \model-H, please refer to M3I Pre-training~\cite{su2022towards}.

\textbf{COCO object detection.} We verify the detection performance of our \model\ on the COCO benchmark~\cite{lin2014microsoft}, on top of Mask R-CNN~\cite{he2017mask} and Cascade Mask R-CNN~\cite{cai2019cascade}.
For fair comparisons, we follow common practices~\cite{liu2021swin,wang2021pvtv2} to initialize the backbone with pre-trained classification weights, and train these models using a 1$\times$ (12 epochs) or 3$\times$ (36 epochs) schedule by default.
For 1$\times$ schedule, the image is resized to have a shorter side of 800 pixels, while the longer side does not exceed 1,333 pixels.
During testing, the shorter side of the input image is fixed to 800 pixels.
For 3$\times$ schedule, the shorter side is resized to 480$-$800 pixels, while the longer side does not exceed 1,333 pixels.
All these detection models are trained with a batch size of 16 and optimized by AdamW~\cite{loshchilov2017decoupled} with an initial learning rate of $1\times10^{-4}$.

\textbf{ADE20K semantic segmentation.} We evaluate our \model\ models on the ADE20K dataset~\cite{zhou2017scene}, and initialize them with the pre-trained classification weights.
For the \model-T/S/B models, we optimize them using AdamW~\cite{loshchilov2017decoupled} with an initial learning rate of 6$\times10^{-5}$, and 2$\times10^{-5}$ for \model-X/XL.
The learning rate is decayed following the polynomial decay schedule with a power of 1.0.
Following previous methods~\cite{liu2021swin,wang2021pvtv2, liu2022convnet}, the crop size is set to 512 for \model-T/S/B, and 640 for \model-L/XL.
All segmentation models are trained using UperNet~\cite{xiao2018unified} with a batch size of 16 for 160k iterations, and compared fairly with previous CNN-based and transformer-based backbones.

\subsection{Settings for System-Level Comparison}

\textbf{COCO object detection.}
For system-level comparison with state-of-the-art large-scale detection models~\cite{wang2022beit3,liu2021swinv2,zhang2022dino,wei2022fdswin,liang2022cbnet}, we first initialize the \model-XL/H backbone with the weights pre-trained on ImageNet-22K or the 427M large-scale joint dataset, and double its parameters using the composite techniques~\cite{liang2022cbnet}.
Then, we pre-train the model along with the DINO~\cite{zhang2022dino} detector on the Objects365~\cite{shao2019objects365} for 26 epochs, with an initial learning rate of $2\times10^{-4}$ and a batch size of 256. 
The shorter size of input images is resized to 600$-$1200 pixels during pre-training, and the learning rate drops by 10 times at epoch 22.
Finally, we fine-tune these detectors on the COCO dataset for 12 epochs, where the batch size is 64, and the initial learning rate is $5\times10^{-5}$, which drops by 10 times at the final epoch.

\textbf{ADE20K semantic segmentation.}
To further reach leading segmentation performance, we first initialize our \model-H backbone with the pre-trained weights on the 427M large-scale joint dataset, and arm it with the state-of-the-art segmentation method Mask2Former \cite{cheng2021masked}. 
We follow the same training settings in \cite{chen2022vitadapter,wang2022beit3}, \ie\ pre-training and fine-tuning the model on COCO-Stuff~\cite{caesar2018coco} and ADE20K~\cite{zhou2017scene} datasets both for 80k iterations, with a crop size of 896 and an initial learning rate of 1$\times10^{-5}$.

\begin{table*}[t]
    \centering
    \renewcommand\arraystretch{1.0}
    \footnotesize
    
\begin{tabular}{@{\ }l|c|c|c|cc|cc|c@{\ }}
\multirow{2}{*}{settings} & \model-T & \model-S & \model-B & \multicolumn{2}{c|}{\model-L} & \multicolumn{2}{c|}{\model-XL} & \model-H \\
\cline{2-9}
& IN-1K pt & 
IN-1K pt &
IN-1K pt & 
IN-22K pt & 
IN-1K ft & IN-22K pt & IN-1K ft & IN-1K ft \\
\hline
input scale & 
224 & 
224 &
224 & 
192 & 
384 & 192 & 384 & 224/640\\
batch size & 
4096 & 
4096 &
4096 & 
4096 & 
512 & 4096 & 512 & 512\\
optimizer &
AdamW & 
AdamW &
AdamW &
AdamW &
AdamW &
AdamW & AdamW & AdamW\\
LR      & 
4$\times10^{-3}$ & 
4$\times10^{-3}$ &
4$\times10^{-3}$ & 
1$\times10^{-3}$ & 
2$\times10^{-5}$ &
1$\times10^{-3}$ & 
2$\times10^{-5}$ & 2$\times10^{-5}$\\
LR schedule& 
cosine  &
cosine & 
cosine & 
cosine & 
cosine &
cosine & cosine & cosine\\
weight decay     &
0.05  & 
0.05  & 
0.05 & 
0.05 &
0.05 &
0.05 & 0.05 & 0.05 \\
warmup epochs & 
5 &
5 &
5 & 
5 &
0 &
5 & 0 & 0 \\
epochs & 
300 &
300 &
300 & 
90 &
20 &
90 & 20 & 20 \\
\hline
horizontal flip  & 
\yes & 
\yes &
\yes & 
\yes & 
\yes  &
\yes  & \yes & \yes
\\
random resized crop & 
\yes & 
\yes & 
\yes & 
\yes & 
\yes & 
\yes & \yes & \yes \\
auto augment  &
\yes & 
\yes &
\yes &
\yes &
\yes &
\yes & \yes & \yes  \\
layer scale  & 
\no  & 
\yes  & 
\yes  &
\yes  &
  \yes&
  \yes &  \yes & \yes\\

mixup alpha  & 
0.8 & 
0.8 & 
0.8 &
0.8 & 
\no  &
0.8 & \no & \no \\
cutmix alpha &
1.0 & 
1.0 & 
1.0 &
1.0 & 
\no &
1.0 & \no & \no \\
erasing prob. &
0.25    &
0.25   &
0.25 &
0.25 &
\no & 
0.25 & \no & \no \\
color jitter  & 
0.4   & 
0.4   & 
0.4  &
0.4  &
0.4 & 
0.4 & 0.4 & 0.4\\

\hline
label smoothing $\varepsilon$ & 
0.1 & 
0.1 &
0.1  &
0.1 & 
0.3  &
0.1  & 0.3 & 0.3 \\%
dropout      & 
\no  & 
\no & 
\no & 
\no  & 
\no &
\no & \no & \no\\
drop path rate & 
0.1 & 
0.4 & 
0.5 & 
0.1 &
0.1 &
0.2 & 0.2 & 0.2 \\
repeated aug & 
\no & 
\no & 
\no &
\no &
\no &
\no & \no & \no\\
gradient clip & 
5.0  & 
5.0 & 
5.0 & 
5.0 & 
5.0 &
5.0 & 5.0 & 5.0\\
loss &
CE & 
CE & 
CE &
CE & 
CE &
CE & CE & CE \\

\end{tabular}
    \caption{\textbf{Detailed training recipe for \model\ of different parameter scales on ImageNet~\cite{deng2009imagenet}.} ``CE'' denotes the cross entropy loss, ``LR'' denotes the learning rate. 
    The training recipe follows common practices~\cite{liu2021swin,touvron2022deit3,touvron2021training,liu2022convnet} and has some tune-ups. ``IN-1K pt'', ``IN-22K pt'', and ``IN-1K ft'' represent ImageNet-1K pre-training, ImageNet-22K pre-training, and ImageNet-1K fine-tuning, respectively.}
    \label{tab:supp_cls_setting}
\end{table*}

\begin{figure}[t]
    \centering
    \includegraphics[width=0.8\columnwidth]{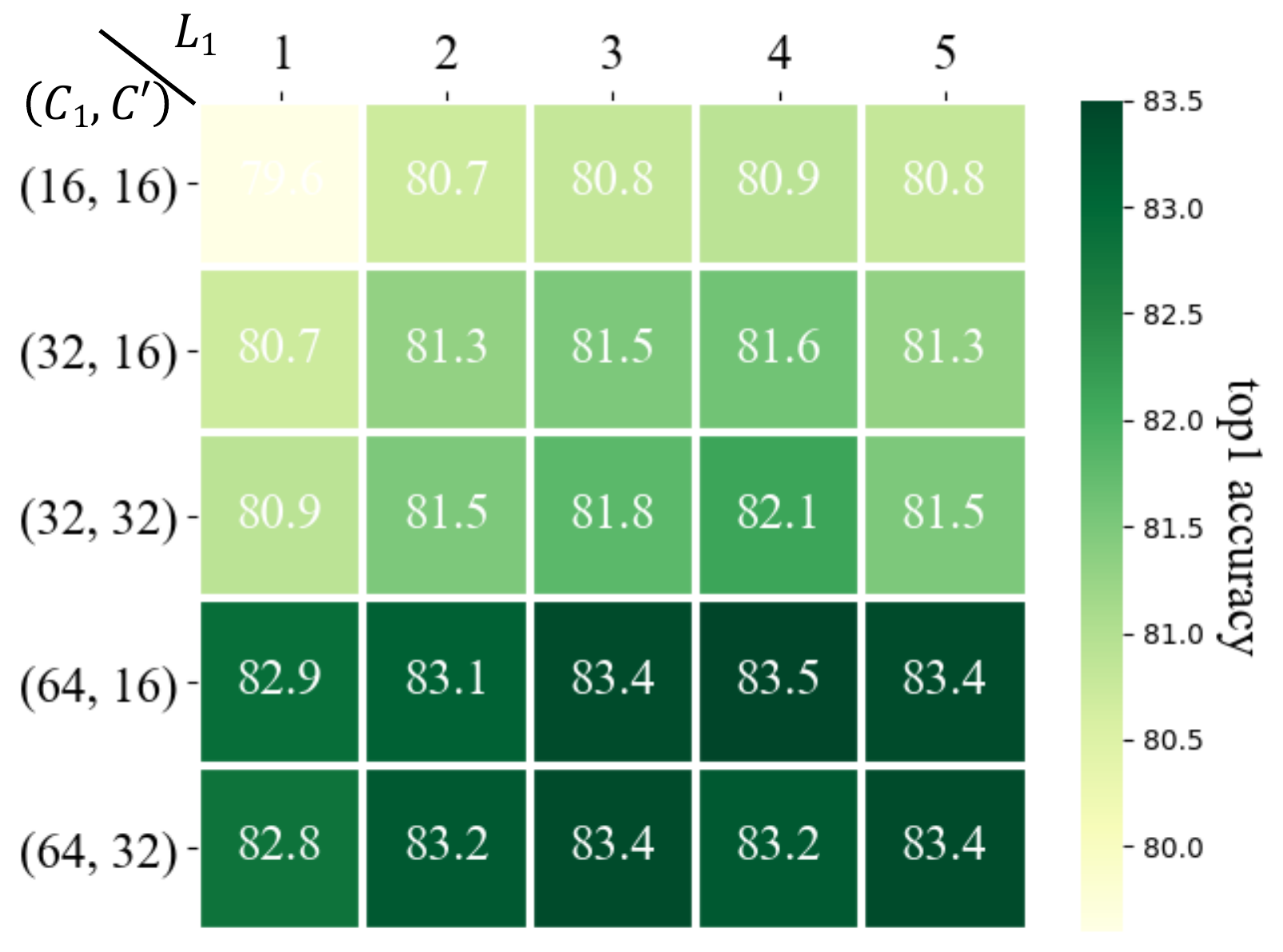}
    \caption{
    \textbf{Comparison of different stacking hyper-parameters}. Each square indicates the accuracy of the model determined by hyperparameter, with the darker the color, the higher the accuracy.
    }
    \label{fig:supp_heatmap}
\end{figure}

\section{Exploration of Hyper-parameters}

\subsection{Model Stacking}
As discussed in Section \textcolor{red}{3.2}, our model is constructed in four stacking rules, and we further restrict the model parameters to 30M for the origin model.
We discretize the stacking hyperparameters $C_1$ to $\{16, 32, 64\}$, $L_1$ to $\{1, 2, 3, 4, 5\}$, and $C'$ to $\{16, 32\}$. And $L_2$ is determined by selecting the model size to approximately 30M. In this way, we obtained 30 models by combining the three hyper-parameters. 

We adopt the training recipe listed in Table~\ref{tab:supp_cls_setting} to train our -T models unless otherwise stated. 
Fig.~\ref{fig:supp_heatmap} shows the ImageNet-1K top-1 accuracy of these models under the same training settings, with darker green indicating higher accuracy, \ie, models with stronger representational capability. 
When $C'$ equals 16, models are generally higher than that with $C'$ of 32, and $L_1$ works best at 4, thanks to a reasonable stacking ratio. A large number of channels allows for more gain.
Finally, through the above exploration experiments, we determine our basic stacking hyper-parameter $(C_1, C', L_1, L_3)$ to $(64, 16, 4, 18)$.

\subsection{Model Scaling}
In Section \textcolor{red}{3.2}, we have shown the constraints on the depth scaling factor $\alpha$ and the width scaling factor $\beta$. 
Based on this condition and the -T model (30M), we display reasonable scaling possibilities for extending the -T model to -B models (100M). 
As illustrated in Table~\ref{tab:supp_scaling_alpha}, the first two columns show the formulas for $\alpha$ and $\beta$. 
The penultimate column indicates model parameters, and the last column indicates the ImageNet-1K top-1 accuracy of these models after 300 training epochs. 

It is worth noting that the model width $C_1$ needs to be divisible by $C'$. Therefore some adjustment is required in determining the specific scaling parameters. 
This results in a small fluctuation in the number of parameters, but this is acceptable.
Our exploratory experiments prove that when $(\alpha, \beta)$ is set at $(1.09, 1.36)$ for the best performance. 
In addition, the other size models -S/L/XL/H also confirmed the effectiveness of our scaling rules.

\begin{table}[t]
    \centering
    \renewcommand\arraystretch{1.0}
    \footnotesize
    \begin{tabular}{cc|c|c}
	\multicolumn{2}{c|}{scaling factors} & \multirow{2}{*}{\#parameters} & \multirow{2}{*}{top-1 accuracy (\%)} \\
        $\alpha$ & $\beta$ & &  \\
	\hline
         1.03 & 1.40 & 118M & 84.5 \\
         1.06 & 1.38 & 95M & 83.8 \\
         \rowcolor{gray!20}
         1.09 & 1.36 & 97M & 84.9 \\
         1.12 & 1.34 & 105M & 83.1 \\
         1.15 & 1.32 & 95M & 81.8 \\
\end{tabular}
    \caption{\textbf{Comparison of different scaling factors}. 
    The default setting is marked with a gray background.
    }
    \label{tab:supp_scaling_alpha}
\end{table}

\begin{table}[t]
    \centering
    \renewcommand\arraystretch{1.0}
    \footnotesize
    \begin{tabular}{c|c|c|c}
    \renewcommand{\arraystretch}{0.1}
    \setlength\tabcolsep{0.97mm}
    kernel size & \#parameters & FLOPs & top-1 accuracy (\%) \\
    \hline
    \rowcolor{gray!20}
    $3\times3$ & 30M & 5G & 83.5 \\
    $5\times5$ & 37M & 6G & 83.6 \\
    $7\times7$ & 48M & 8G & 82.8 \\
\end{tabular}
    \caption{\textbf{Comparison of different kernel sizes in our operator}.
    The default setting is marked with a gray background.}
    \label{tab:supp_large_kernel}
\end{table}
\begin{figure}[t]
    \centering
    \includegraphics[width=0.9\columnwidth]{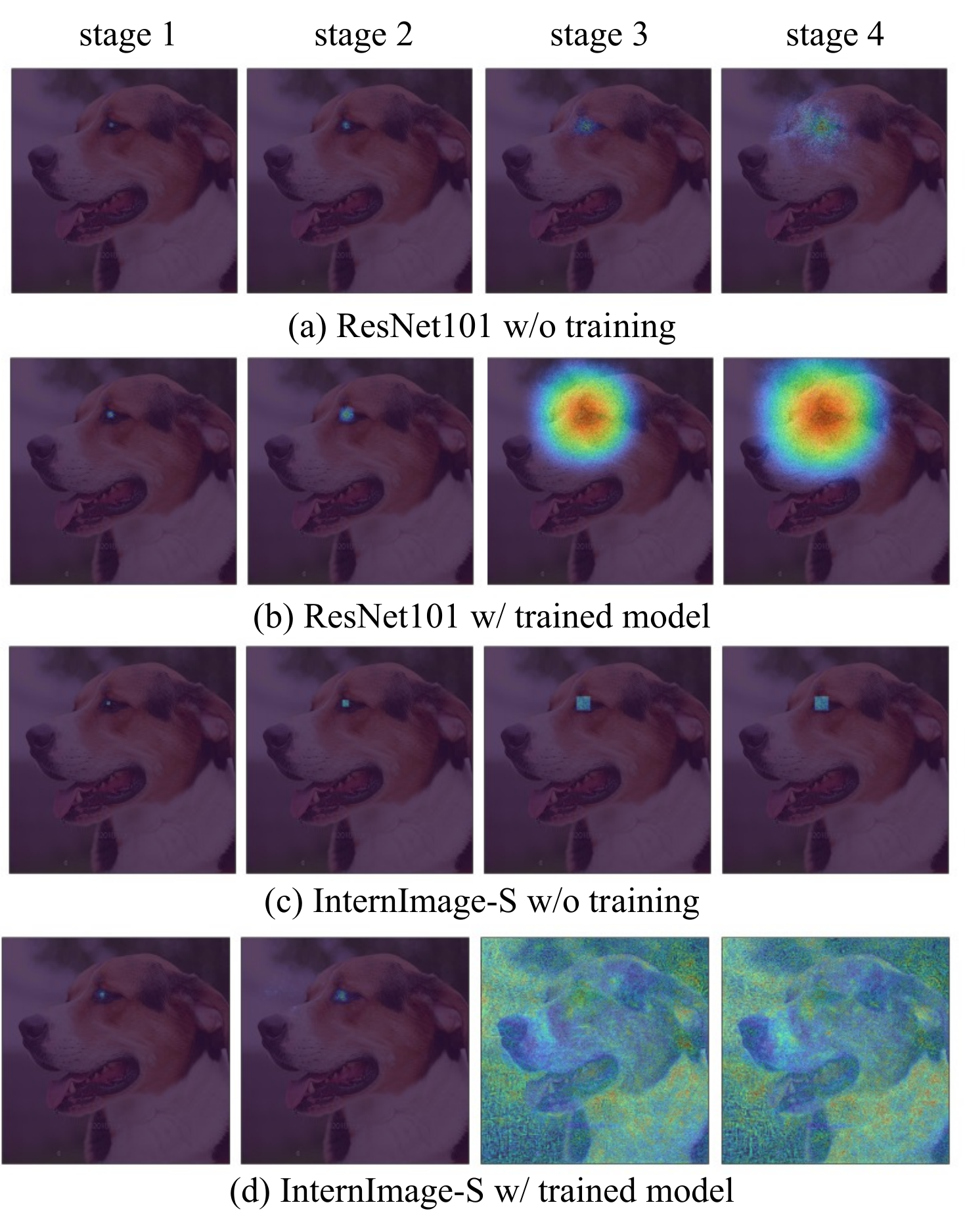}
    \caption{
    \textbf{Visualization of the effective receptive field (ERF) of different backbones.} The activated pixel is at dog's eye.
    (a) and (b) shows the ERF of ResNet-101~\cite{he2016deep} with (w/) and without (w/o) training on ImageNet-1K~\cite{deng2009imagenet}, respectively. (c) and (d) are the ERF of \model-B with (w/) and without (w/o) training on ImageNet-1K. 
    }
    \label{fig:supp_erf}
\end{figure}

\subsection{Kernel Size}
As mentioned in Section \textcolor{red}{3.1}, we argue 3$\times$3 dynamic sparse convolution is enough for the large receptive field. 
Here, we explore the role played by the number of convolutional neurons in the DCNv3 operator.
Specifically, we replaced the $3\times3$ kernel in the DCNv3 operator with the $5\times5$ or $7\times7$ kernel.
They are all trained by the -T training recipes (see Table~\ref{tab:supp_cls_setting}) and validated on the ImageNet-1K validation set. The results are shown in Table~\ref{tab:supp_large_kernel}.

The results show that when enlarging the convolution kernel, the parameters and FLOPs are followed by the surge, while the accuracy is not significantly improved (83.5 \textit{v.s} 83.6) or even decreased (83.5 \textit{v.s} 82.8). These results show that when the number of convolutional neurons in a single layer increases, the model becomes more difficult to optimize. This phenomenon is also confirmed in RepLKNet~\cite{ding2022replknet}, and it addresses this problem by re-parameterizing~\cite{ding2022replknet} techniques, which might bring extra time and memory costs in the training phase.
In this work, we avoid this problem by adopting the simple yet effective $3\times3$ DCNv3 as \model's core operator.

Fig.~\ref{fig:supp_erf} shows the effective receptive fields (ERF) of ResNet-101~\cite{he2016deep} and \model-S. 
A wider distribution of bright areas indicates a larger ERF. 
We uniformly activate the input image at the dog's eye, count the gradient map of each block, aggregate by channel, and map back to the input image.
We see that the ERF of ResNet-101~\cite{he2016deep} without training is limited to a local area, while the fully trained ResNet-101 still has an ERF around the eye, and the gradient amplitude is lower, and the distribution is more sparse. 
Therefore, the area that ResNet-101 can effectively perceive is very limited.
For the \model-S without training, its ERF is concentrated around the activation point.
Since the offset is not learned, its ERF is also very small in the last two blocks. But after sufficient training, \model-L can effectively perceive the information of the entire image in the 3-rd and 4-th stages.

\begin{table*}[t]
    \centering
    \renewcommand\arraystretch{1.0}
    \setlength{\tabcolsep}{1.7mm}
    \footnotesize
    \setlength\tabcolsep{1mm}
\begin{tabular}{l|lll|lllll}
    \renewcommand{\arraystretch}{0.1}
    \multirow{2}{*}{method} & \multicolumn{3}{c|}{classification} & \multicolumn{5}{c}{semantic segmentation} \\
                  & iNaturalist2018             & Places205 & Places365 & COCO-Stuff-10K & Pascal Context & Cityscapes (val) & Cityscapes (test) & NYU Depth V2 \\
    \hline 
    previous best & 88.7$^{\rm a}$       & 69.3$^{\rm b}$ & 60.7$^{\rm c}$ & 54.2$^{\rm d}$ & 68.2$^{\rm d}$ & 86.9$^{\rm e}$ & 85.2$^{\rm d}$ & 56.9$^{\rm f}$\\
    InternImage-H &  92.6 \green{(+3.9)} & 71.7 \green{(+2.4)} & 61.2 \green{(+0.5)} & 59.6 \green{(+5.4)} & 70.3 \green{(+2.1)} & 87.0 \green{(+0.1)} & 86.1 \green{(+0.9)} & 68.1 \green{(+11.2)} \\
\end{tabular}

\setlength\tabcolsep{3.5mm}
\begin{tabular}{l|lllllll}
    \renewcommand{\arraystretch}{0.1}
    \multirow{2}{*}{method} &  \multicolumn{7}{c}{object detection}  \\
     &  LVIS (minival) & LVIS (val) & VOC2007 & VOC2012 & OpenImages & CrowdHuman & BDD100K   \\
    \hline
    previous best  & 59.8$^{\rm g}$ & 62.2$^{\rm h}$ & 89.3$^{\rm i}$ & 92.9$^{\rm j}$ & 72.2$^{\rm k}$ & 94.1$^{\rm l}$ & 35.6$^{\rm m}$ \\
    InternImage-H  & 65.8 \green{(+6.0)} & 63.2 \green{(+1.0)} & 94.0 \green{(+4.7)} & 97.2 \green{(+4.3)} & 74.1 \green{(+1.9)} & 97.2 \green{(+3.1)} & 38.8 \green{(+3.2)}  \\
\end{tabular}
    \caption{\textbf{Summary of InternImage-H performance on various mainstream vision benchmarks}.
    a: MetaFormer~\cite{diao2022metaformer}. 
    b: MixMIM-L~\cite{liu2022mixmim}.
    c: SWAG~\cite{singh2022revisiting}.
    d: ViT-Adapter~\cite{chen2022vitadapter}.
    e: PSA~\cite{liu2021polarized}.
    f: CMX-B5~\cite{liu2022cmx}.
    g: GLIPv2~\cite{zhang2022glipv2}.
    h: EVA~\cite{fang2022eva}.
    i: Cascade Eff-B7 NAS-FPN~\cite{ghiasi2021simple}.
    j: ATLDETv2~\cite{jin2016ATLDETv2}.
    k: OpenImages 2019 competition 1$^{\rm st}$~\cite{liu20201st}.
    l: Iter-Deformable-DETR~\cite{zheng2022progressive}.
    m: PP-YOLOE~\cite{xu2022pp}.
    }
    \label{tab:supp_downstrem_sota}
\end{table*}

\section{Additional Downstream Tasks}

\subsection{Classification}

\textbf{iNaturalist 2018}~\cite{van2018inaturalist} is a read-word long-tailed dataset containing 8142 fine-graned species. The dataset comprises 437.5K training images and an imbalance factor of 500. 
For this experiment, we initialize our \model-H model with the pre-trained weights on the 427M large-scale joint dataset, and fine-tune it on the training set of iNaturalist 2018 for 100 epochs. 
We follow MetaFormer \cite{diao2022metaformer} to adopt a resolution of 384$\times$384 for fine-tuning, with the utilization of meta information.
Other training settings are the same as the recipe for fine-tuning \model-H on ImageNet-1K, as reported in Table~\ref{tab:supp_cls_setting}.
As a result, our method achieves the state-of-the-art accuracy of 92.6 (see Table~\ref{tab:supp_downstrem_sota}) on the validation set of iNaturalist 2018, 3.9 points better than the previous best model MetaFormer~\cite{diao2022metaformer}.

\textbf{Places205}~\cite{zhou2014learning} is a dataset containing 2.5 million images of 205 scene categories, which are dedicated to the scene recognition task. The images in this dataset cover a wide range of indoor and outdoor scenes, such as offices, kitchens, forests, and beaches. We initialize our model with pre-trained weights on a large-scale joint dataset, consisting of 427 million images, and fine-tune it on the Places205 training set. Other training settings are the same as the recipe for fine-tuning \model-H on ImageNet-1K, as reported in Table~\ref{tab:supp_cls_setting}. Our method achieves state-of-the-art accuracy of 71.7 (see Table~\ref{tab:supp_downstrem_sota}) on the validation set of Places205, outperforming the previous best model MixMIM-L~\cite{liu2022mixmim} by 2.4 points.

\textbf{Places365}~\cite{lopez2020semantic} is a dataset containing 1.8 million images of 365 scene categories, which are dedicated to the scene recognition task. The images in this dataset cover a wide range of indoor and outdoor scenes, such as airports, bedrooms, deserts, and waterfalls. The specific pre-training and fine-tuning strategies are the same as for Places205. Our method achieves state-of-the-art accuracy of 61.2 (see Table~\ref{tab:supp_downstrem_sota}) on the validation set of Places365, outperforming the previous best model SWAG~\cite{singh2022revisiting} by 0.5 points. The Places365 dataset provides a more fine-grained classification task compared to Places205, allowing our model to learn more subtle differences between similar scenes.

\subsection{Object Detection}

\textbf{LVIS v1.0}~\cite{gupta2019lvis} is a large-scale vocabulary dataset for object detection and instance segmentation tasks, which contains 1203 categories in 164k images. For this dataset, we initialize our \model-H with the Objects365~\cite{shao2019objects365} pre-trained weights, then fine-tune it on the training set of LVIS v1.0.
Here, we report the box AP (\ie, AP${\rm ^b}$) with multi-scale testing on the minival set and the val set, respectively.
As shown in Table~\ref{tab:supp_downstrem_sota}, our \model-H creates a new record of 65.8 AP${\rm ^b}$ on the LVIS minival, and 63.2 AP${\rm ^b}$ on the LVIS val, outperforming previous state-of-the-art methods by clear margins.

\textbf{Pascal VOC}~\cite{everingham2010pascal} contains 20 object classes, which has been widely used as a benchmark for object detection tasks. 
We adopt this dataset to further evaluate the detection performance of our model.
Specifically, we employ the Objects365~\cite{shao2019objects365} pre-trained weights to initialize our \model-H, and fine-tune it on the trainval set of Pascal VOC 2007 and Pascal VOC 2012 following previous method~\cite{ghiasi2021simple}.
As shown in Table~\ref{tab:supp_downstrem_sota}, on the Pascal VOC 2007 test set, our \model-H yields 94.0 AP$^{\rm 50}$ with single-scale testing, which is 4.7 points better than previous best Cascade Eff-B7 NAS-FPN~\cite{ghiasi2021simple}.
On the Pascal VOC 2012 test set, our method achieves 97.2 mAP, 4.3 points higher than the best record on the official leaderboard~\cite{jin2016ATLDETv2}.

\textbf{OpenImages v6}~\cite{kuznetsova2020open} is a dataset of about 9 million images with 16M bounding boxes for 600 object classes on 1.9 million images dedicated to the object detection task, which are very diverse and often embrace complex scenes with multiple objects (8.3 per image on average).
For this dataset, we use the same settings as the previous two datasets. In addition, we follow \cite{liu20201st} to use the class-aware sampling during fine-tuning.
As reported in Table~\ref{tab:supp_downstrem_sota}, our \model-H yields 74.1 mAP, achieving 1.9 mAP improvement compared to the previous best results~\cite{liu20201st}.

\textbf{CrownHuman}~\cite{shao2018crowdhuman} is a benchmark dataset to better evaluate detectors in crowd scenarios. The CrowdHuman dataset is large, rich-annotated and contains high diversity. CrowdHuman contains 15000, 4370 and 5000 images for training, validation, and testing, respectively. There are a total of 470K human instances from train and validation subsets and 23 persons per image, with various kinds of occlusions in the dataset. We used the same training setup as for the previous dataset. Our pre-trained model reached optimal performance in 3750 iterations, exceeding the previous best model Iter-Deformable-DETR~\cite{zheng2022progressive} by 3.1 AP.

\textbf{BDD100K}~\cite{yu2020bdd100k} is a dataset of around 100K high-resolution images with diverse weather and lighting conditions, containing 10 object categories, including pedestrians, cars, buses, and bicycles, dedicated to the object detection task. The images in this dataset are captured from a moving vehicle, simulating real-world scenarios. 
For this experiment, we initialize our \model-H model with the pre-trained weights on the 427M joint dataset and fine-tune it on the BDD100K training set for 12 epochs. 
As reported in Table~\ref{tab:supp_downstrem_sota}, our \model-H achieves 38.8 mAP on the validation set, which is the state-of-the-art performance, surpassing the previous best model by 3.2 mAP. Our method demonstrates superior performance in detecting objects in real-world driving scenarios, which can benefit autonomous driving and intelligent transportation systems.

\subsection{Semantic Segmentation}

\textbf{COCO-Stuff} \cite{caesar2018coco} includes the images from the COCO \cite{lin2014microsoft} dataset for semantic segmentation, spanning over 171 categories.
Specifically, COCO-Stuff-164K is the full set that contains all 164k images, while COCO-Stuff-10K is a subset of the -164K that splits into 9,000 and 1,000 images for training and testing. 
Here, we equip our \model-H with the advanced Mask2Former~\cite{cheng2021masked}, and pre-train the model on the COCO-Stuff-164K for 80k iterations.
Then we fine-tune it on the COCO-Stuff-10K for 40k iterations and report the multi-scale mIoU.
The crop size is set to 512$\times$512 in this experiment.
As shown in Table~\ref{tab:supp_downstrem_sota}, our model achieves 59.6 MS mIoU on the test set, outperforming the previous best ViT-Adapter~\cite{chen2022vitadapter} by 5.4 mIoU.

\textbf{Pascal Context}~\cite{mottaghi2014role} contains 59 semantic classes. It is divided into 4,996 images for training and 5,104 images for testing.
For this dataset, we also employ Mask2Former with our \model-H, and follow the training settings in \cite{chen2022vitadapter}. 
Specifically, we first load the classification pre-trained weights to initialize the model, then fine-tune it on the training set of Pascal Context for 40k iterations. 
The crop size is set to 480$\times$480 in this experiment.
As shown in Table~\ref{tab:supp_downstrem_sota}, our method reports 70.3 MS mIoU on the test set, which is 2.1 points better than ViT-Adapter~\cite{chen2022vitadapter}.

\textbf{Cityscapes}~\cite{Cordts_2016_CVPR} is a high-resolution dataset recorded in street scenes including 19 classes. 
In this experiment, we use Mask2Former~\cite{cheng2021masked} as the segmentation framework. 
Following common practices~\cite{tao2020hierarchical,xie2021segformer,chen2022vitadapter}, we first pre-train on Mapillary Vistas~\cite{neuhold2017mapillary} and then fine-tune on Cityscapes for 80k iterations, respectively.
The crop size is set to 1024$\times$1024 in this experiment.
As shown in Table~\ref{tab:supp_downstrem_sota}, our \model-H achieves 87.0 MS mIoU on the validation set, and 86.1 MS mIoU on the test set.

\textbf{NYU Depth V2}~\cite{silberman2012indoor} comprises of 1449 RGB-D images, each with a size of 640$\times$480. These images are divided into 795 training and 654 testing images, each with annotations on 40 semantic categories. 
We adopt the same training settings as we used when fine-tuning on Pascal Context. 
As shown in Table~\ref{tab:supp_downstrem_sota}, our method achieves a big jump to 68.1 MS mIoU on the validation set, which is 11.2 points better than CMX-B5~\cite{liu2022cmx}.

\begin{table}[t]
    \centering
    \renewcommand\arraystretch{1.0}
    \setlength{\tabcolsep}{0.5mm}
    \footnotesize
    \begin{tabular}{l|c|c|c|c|c}
    \renewcommand{\arraystretch}{0.1}
    method & \#params & scale & FLOPs & acc (\%) & throughput (img/s) \\
    \hline
    \rowcolor{gray!20}
    \model-B  & & 224$^2$ & 16G & 84.9 & 775 \\
    \rowcolor{gray!20}
    (ours) & \multirow{-2}{*}{97M} & 800$^2$ & 206G & $-$ &  54 \\
     \hline
    \model-B-  & & 224$^2$ & 24G & $-$ & 311 \\
    DCNv2~\cite{zhu2019dcnv2}& \multirow{-2}{*}{146M} & 800$^2$ & 313G & $-$ & 16 \\
    \hline
    \multirow{2}{*}{ConvNeXt-B~\cite{liu2022convnet}} & \multirow{2}{*}{88M} & 224$^2$ &  15G & 83.8  & 881 \\
    & & 800$^2$ & 196G & $-$ & 58 \\
    \hline
    \multirow{2}{*}{RepLKNet-B~\cite{ding2022replknet}} & \multirow{2}{*}{79M} & 224$^2$ & 15G & 83.5 & 884 \\
    & & 800$^2$ & 198G & $-$ & 21 \\
    \hline
    \multirow{2}{*}{DAT-B~\cite{liu2022convnet}} & \multirow{2}{*}{88M} & 224$^2$ & 16G & 84.0 & 661 \\
    & & 800$^2$ & 194G & $-$ & 24 \\
\end{tabular}
    \caption{\textbf{Throughput comparison of different models under different input resolutions.} ``\#params'' denotes the number of parameters. ``acc'' represents the top-1 accuracy on the ImageNet-1K validation set. The throughputs of 224$\times$224 and 800$\times$800 input resolutions are tested with the batch size of 256 and 2 respectively, using a single A100 GPU.}
    \label{tab:speed_cmp}
\end{table}

\section{Throughput Analysis}

In this section, we benchmark the throughput of our \model\ with counterparts, including a variant equipped with DCNv2~\cite{zhu2019dcnv2}, ConvNext~\cite{liu2022convnet}, RepLKNet~\cite{ding2022replknet}, and a vision transformer with deformable attention (DAT)~\cite{xia2022vision}.
As shown in Table~\ref{tab:speed_cmp}, compared to the variant with DCNv2 \cite{zhu2019dcnv2}, our model enjoys better parameter-efficient and significantly faster inference speed under both 224$\times$224 and 800$\times$800 input resolutions.
Compared to RepLKNet-B~\cite{ding2022replknet} and DAT-B~\cite{xia2022vision}, our model has a throughput advantage at a high input resolution (\ie, 800$\times$800). This resolution is widely used in dense prediction tasks such as object detection.
Compared with ConvNeXt~\cite{liu2022convnet}, despite the throughput gap due to DCN-based operators, our model still has an accuracy advantage (84.9 \vs 83.8), and we are also looking for an efficient DCN to make our model more suitable for downstream tasks that require high efficiency.

\section{Robustness Evaluation on ImageNet}
\begin{figure}[t]
    \centering
    \includegraphics[width=1.0\columnwidth]{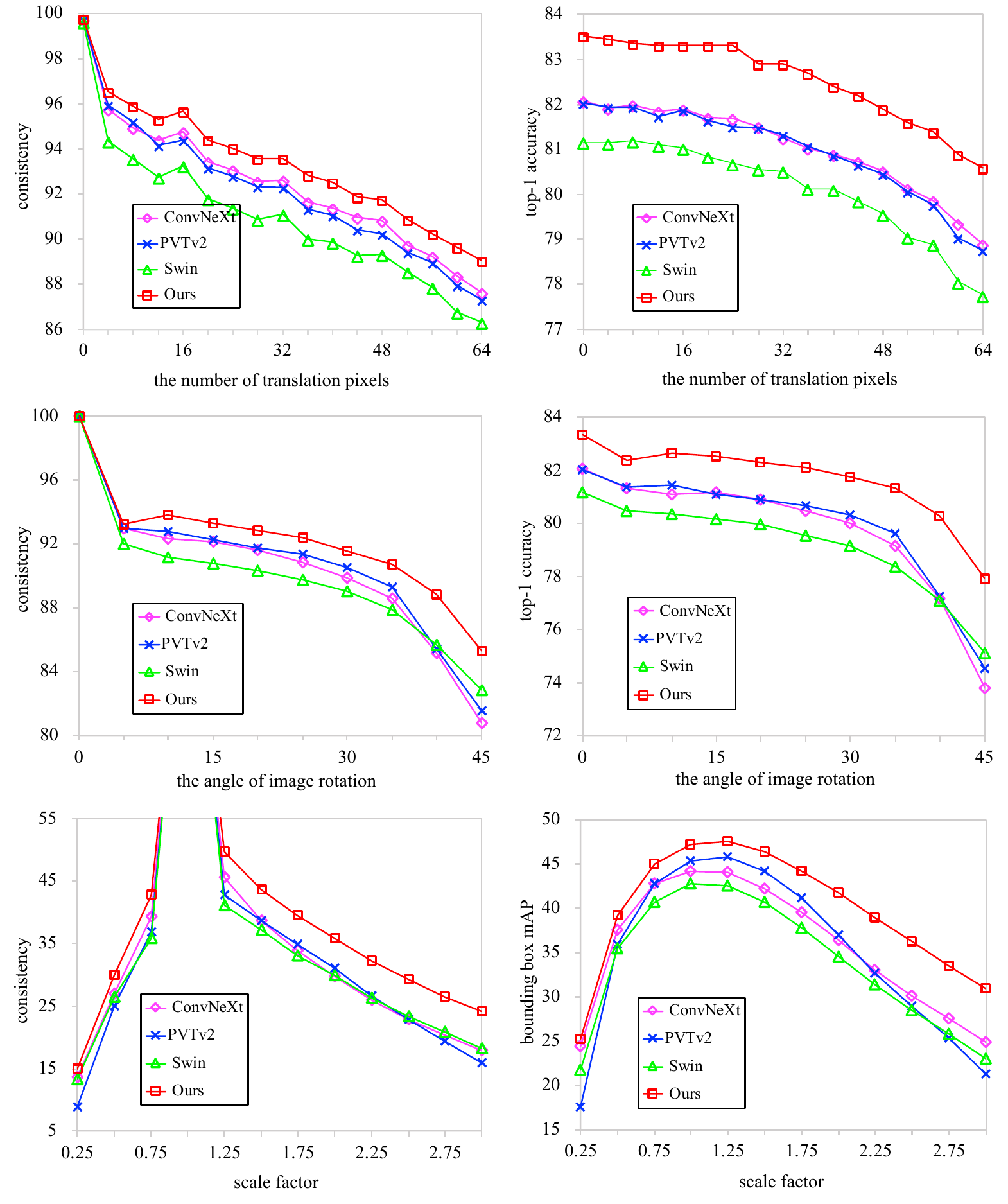}
    \caption{
    \textbf{Comparison of robust evaluation of different methods. }
    These results show that our model has better robustness in terms of translation, rotation, and input resolution.
    }
    \label{fig:supp_robust}
\end{figure}

In this section, we evaluate the robustness of different models under different transformations (see Fig.~\ref{fig:supp_robust}).
We consider translation, rotation, and scaling to evaluate. 
The models we choose for comparison include a convolutional model (ConvNeXt-T~\cite{liu2022convnet}), a local attention-based model (Swin-T~\cite{liu2021swin}), a global attention-based model (PVTv2-B2~\cite{wang2021pvtv2}), and our \model-T.

\subsection{Translation Invariance}
Translation invariance describes the capability of the model to retain the original output when the input image is translated. 
We evaluate the translation invariance in the classification task by dithering the image from 0 to 64 pixels. 
The invariance is measured by the probability that the model predicts the same label when the same input image is translated.
The first row of Fig.~\ref{fig:supp_robust} indicates our \model has the translation invariance of the different methods. 
It is evident that the robustness of the four models to translation is shown as our method is the best, followed by convolution-based ConvNeXt, followed by global attention-based PVTv2, and the worst local attention-based Swin transformer.

\subsection{Rotation Invariance}
To evaluate the rotation invariance of the classification task, we rotate the image from $0\degree$ to $45\degree$ in steps of $5\degree$.
In a similar way to translation invariance, the predicted consistency under different rotation angles is used to evaluate the rotational invariance.
From the second row of Fig.~\ref{fig:supp_robust}, we found that the consistency performance of all models is comparable in the small angle phase. 
However, at large-angle rotation (\ie, $>10\degree$), our model is clearly superior to the other models. 

\subsection{Scaling Invariance}
We evaluate the scaling invariance on object detection. 
The scaling factor of the input image varies from 0.25 to 3.0 in steps of 0.25. 
Detection consistency is defined as the invariance metric for the detection task. 
The predicted boxes on the scaled images are first converted back to the original resolution, and then the predicted boxes at the original resolution are used as the ground truth boxes to calculate the box mAP.
As seen in the last row of Fig.~\ref{fig:supp_robust}, we can observe that all methods of our experiments are sensitive to down-scaling. And they show invariance comparable to the input at small resolutions. Our method performs better when scaling up the images. Both box consistency and bounding box mAP are better than the others.

\subsection{How Hungry the Model is for Data Scale?}
In order to verify the robustness of the model to the data scale. We uniformly sampled the ImageNet-1K data to obtain 1\%, 10\%, and 100\% data, respectively. 
And we chose ResNet-50~\cite{he2016deep}, ConvNeXt-T~\cite{liu2022convnet}, Swin-T~\cite{liu2021swin}, \model-T-dattn and our \model-T to conduct 300 rounds of training experiments on these data. The experimental settings are consistent with Table~\ref{tab:supp_cls_setting}.
The experimental results can be viewed in Table~\ref{tab:supp_data_scale}. We see that ResNet~\cite{he2016deep} performs best on the 1\% and 10\% data (12.2\% \& 57.5\%), benefiting from its inductive biases. But its upper limitation is low (80.4\%) when the data is sufficient. 
Swin-T fails completely in 1\% datasets and shows good performance only on the 100\% dataset.
The proposed \model-T has strong robustness not only on 1\% and 10\% data (5.9\% and 56.0\%) but also on full data (83.5\%), which is consistently better than the \model-T variant with deformable attention (dattn) and ConvNeXt~\cite{liu2022convnet}. These results indicate the robustness of our model with respect to the data scale.

\begin{table}[t]
    \centering
    \renewcommand\arraystretch{1.0}
    \footnotesize
    \begin{tabular}{l|c|c|c}
    \renewcommand{\arraystretch}{0.1}
    \setlength\tabcolsep{0.97mm}
    method & 1\% & 10\% & 100\% \\
    \hline
    ResNet-50~\cite{he2016deep} & 12.2 & 57.5 & 80.4 \\
    ConvNeXt-T~\cite{liu2022convnet} & 8.4 & 52.6 & 82.1 \\
    Swin-T~\cite{liu2021swin} & failed & 12.1 & 81.3 \\
    \model-T-dattn~\cite{zhu2020deformabledetr} & 4.1 & 49.9 & 81.9 \\
    \rowcolor{gray!20}
    \model-T (ours) & 5.9 & 56.0 & 83.5 \\
\end{tabular}
    \caption{\textbf{Accuracy of different models at different data scales}. ``\model-dattn'' refers to the model variant equipped with deformable attention~\cite{zhu2020deformabledetr}.}
    \label{tab:supp_data_scale}
\end{table}

{\small
\bibliographystyle{unsrt}
\bibliography{egbib}

\begin{thebibliography}{100}

\bibitem{vaswani2017attention}
Ashish Vaswani, Noam Shazeer, Niki Parmar, Jakob Uszkoreit, Llion Jones,
  Aidan~N Gomez, {\L}ukasz Kaiser, and Illia Polosukhin.
\newblock Attention is all you need.
\newblock {\em Adv. Neural Inform. Process. Syst.}, 30, 2017.

\bibitem{liu2021swin}
Ze~Liu, Yutong Lin, Yue Cao, Han Hu, Yixuan Wei, Zheng Zhang, Stephen Lin, and
  Baining Guo.
\newblock Swin transformer: Hierarchical vision transformer using shifted
  windows.
\newblock In {\em Int. Conf. Comput. Vis.}, pages 10012--10022, 2021.

\bibitem{shoeybi2019megatron}
Mohammad Shoeybi, Mostofa Patwary, Raul Puri, Patrick LeGresley, Jared Casper,
  and Bryan Catanzaro.
\newblock Megatron-lm: Training multi-billion parameter language models using
  model parallelism.
\newblock {\em arXiv preprint arXiv:1909.08053}, 2019.

\bibitem{radford2019gpt2}
Alec Radford, Jeffrey Wu, Rewon Child, David Luan, Dario Amodei, Ilya
  Sutskever, et~al.
\newblock Language models are unsupervised multitask learners.
\newblock {\em OpenAI blog}, 1(8):9, 2019.

\bibitem{raffel2020t5}
Colin Raffel, Noam Shazeer, Adam Roberts, Katherine Lee, Sharan Narang, Michael
  Matena, Yanqi Zhou, Wei Li, and Peter~J Liu.
\newblock Exploring the limits of transfer learning with a unified text-to-text
  transformer.
\newblock {\em Journal of Machine Learning Research}, 21:1--67, 2020.

\bibitem{brown2020gpt3}
Tom Brown, Benjamin Mann, Nick Ryder, Melanie Subbiah, Jared~D Kaplan, Prafulla
  Dhariwal, Arvind Neelakantan, Pranav Shyam, Girish Sastry, Amanda Askell,
  et~al.
\newblock Language models are few-shot learners.
\newblock {\em Adv. Neural Inform. Process. Syst.}, 33:1877--1901, 2020.

\bibitem{chowdhery2022palm}
Aakanksha Chowdhery, Sharan Narang, Jacob Devlin, Maarten Bosma, Gaurav Mishra,
  Adam Roberts, Paul Barham, Hyung~Won Chung, Charles Sutton, Sebastian
  Gehrmann, et~al.
\newblock Palm: Scaling language modeling with pathways.
\newblock {\em arXiv preprint arXiv:2204.02311}, 2022.

\bibitem{fedus2022switch}
William Fedus, Barret Zoph, and Noam Shazeer.
\newblock Switch transformers: Scaling to trillion parameter models with simple
  and efficient sparsity.
\newblock {\em Journal of Machine Learning Research}, 23(120):1--39, 2022.

\bibitem{dosovitskiy2020image}
Alexey Dosovitskiy, Lucas Beyer, Alexander Kolesnikov, Dirk Weissenborn,
  Xiaohua Zhai, Thomas Unterthiner, Mostafa Dehghani, Matthias Minderer, Georg
  Heigold, Sylvain Gelly, et~al.
\newblock An image is worth 16x16 words: Transformers for image recognition at
  scale.
\newblock In {\em Int. Conf. Learn. Represent.}, 2020.

\bibitem{wang2021pyramid}
Wenhai Wang, Enze Xie, Xiang Li, Deng-Ping Fan, Kaitao Song, Ding Liang, Tong
  Lu, Ping Luo, and Ling Shao.
\newblock Pyramid vision transformer: A versatile backbone for dense prediction
  without convolutions.
\newblock In {\em Int. Conf. Comput. Vis.}, pages 568--578, 2021.

\bibitem{wang2021pvtv2}
Wenhai Wang, Enze Xie, Xiang Li, Deng-Ping Fan, Kaitao Song, Ding Liang, Tong
  Lu, Ping Luo, and Ling Shao.
\newblock Pvt v2: Improved baselines with pyramid vision transformer.
\newblock {\em Computational Visual Media}, 8(3):415--424, 2022.

\bibitem{dong2021cswin}
Xiaoyi Dong, Jianmin Bao, Dongdong Chen, Weiming Zhang, Nenghai Yu, Lu~Yuan,
  Dong Chen, and Baining Guo.
\newblock Cswin transformer: A general vision transformer backbone with
  cross-shaped windows.
\newblock {\em IEEE Conf. Comput. Vis. Pattern Recog.}, pages 12124--12134,
  2022.

\bibitem{wu2021cvt}
Haiping Wu, Bin Xiao, Noel Codella, Mengchen Liu, Xiyang Dai, Lu~Yuan, and Lei
  Zhang.
\newblock Cvt: Introducing convolutions to vision transformers.
\newblock In {\em Int. Conf. Comput. Vis.}, pages 22--31, 2021.

\bibitem{ali2021xcit}
Alaaeldin Ali, Hugo Touvron, Mathilde Caron, Piotr Bojanowski, Matthijs Douze,
  Armand Joulin, Ivan Laptev, Natalia Neverova, Gabriel Synnaeve, Jakob
  Verbeek, et~al.
\newblock Xcit: Cross-covariance image transformers.
\newblock {\em Adv. Neural Inform. Process. Syst.}, 34, 2021.

\bibitem{han2021transformer}
Kai Han, An~Xiao, Enhua Wu, Jianyuan Guo, Chunjing Xu, and Yunhe Wang.
\newblock Transformer in transformer.
\newblock {\em Adv. Neural Inform. Process. Syst.}, 34, 2021.

\bibitem{liu2021swinv2}
Ze~Liu, Han Hu, Yutong Lin, Zhuliang Yao, Zhenda Xie, Yixuan Wei, Jia Ning, Yue
  Cao, Zheng Zhang, Li~Dong, et~al.
\newblock Swin transformer v2: Scaling up capacity and resolution.
\newblock {\em Adv. Neural Inform. Process. Syst.}, pages 12009--12019, 2022.

\bibitem{wang2022beit3}
Wenhui Wang, Hangbo Bao, Li~Dong, Johan Bjorck, Zhiliang Peng, Qiang Liu, Kriti
  Aggarwal, Owais~Khan Mohammed, Saksham Singhal, Subhojit Som, et~al.
\newblock Image as a foreign language: Beit pretraining for all vision and
  vision-language tasks.
\newblock {\em arXiv preprint arXiv:2208.10442}, 2022.

\bibitem{riquelme2021vmoe}
Carlos Riquelme, Joan Puigcerver, Basil Mustafa, Maxim Neumann, Rodolphe
  Jenatton, Andr{\'e} Susano~Pinto, Daniel Keysers, and Neil Houlsby.
\newblock Scaling vision with sparse mixture of experts.
\newblock {\em Adv. Neural Inform. Process. Syst.}, 34:8583--8595, 2021.

\bibitem{zhai2022scalingvit}
Xiaohua Zhai, Alexander Kolesnikov, Neil Houlsby, and Lucas Beyer.
\newblock Scaling vision transformers.
\newblock In {\em IEEE Conf. Comput. Vis. Pattern Recog.}, pages 12104--12113,
  2022.

\bibitem{dai2021coatnet}
Zihang Dai, Hanxiao Liu, Quoc~V Le, and Mingxing Tan.
\newblock Coatnet: Marrying convolution and attention for all data sizes.
\newblock {\em Adv. Neural Inform. Process. Syst.}, 34:3965--3977, 2021.

\bibitem{liu2022convnet}
Zhuang Liu, Hanzi Mao, Chao-Yuan Wu, Christoph Feichtenhofer, Trevor Darrell,
  and Saining Xie.
\newblock A convnet for the 2020s.
\newblock {\em arXiv preprint arXiv:2201.03545}, 2022.

\bibitem{ding2022replknet}
Xiaohan Ding, Xiangyu Zhang, Jungong Han, and Guiguang Ding.
\newblock Scaling up your kernels to 31x31: Revisiting large kernel design in
  cnns.
\newblock In {\em IEEE Conf. Comput. Vis. Pattern Recog.}, pages 11963--11975,
  2022.

\bibitem{yu2022metaformer}
Weihao Yu, Mi~Luo, Pan Zhou, Chenyang Si, Yichen Zhou, Xinchao Wang, Jiashi
  Feng, and Shuicheng Yan.
\newblock Metaformer is actually what you need for vision.
\newblock In {\em IEEE Conf. Comput. Vis. Pattern Recog.}, pages 10819--10829,
  2022.

\bibitem{ba2016layernorm}
Jimmy~Lei Ba, Jamie~Ryan Kiros, and Geoffrey~E Hinton.
\newblock Layer normalization.
\newblock {\em arXiv preprint arXiv:1607.06450}, 2016.

\bibitem{hendrycks2016gelu}
Dan Hendrycks and Kevin Gimpel.
\newblock Gaussian error linear units (gelus). arxiv.
\newblock {\em arXiv preprint arXiv:1606.08415}, 2016.

\bibitem{wei2022fdswin}
Yixuan Wei, Han Hu, Zhenda Xie, Zheng Zhang, Yue Cao, Jianmin Bao, Dong Chen,
  and Baining Guo.
\newblock Contrastive learning rivals masked image modeling in fine-tuning via
  feature distillation.
\newblock {\em arXiv preprint arXiv:2205.14141}, 2022.

\bibitem{dai2017dcnv1}
Jifeng Dai, Haozhi Qi, Yuwen Xiong, Yi~Li, Guodong Zhang, Han Hu, and Yichen
  Wei.
\newblock Deformable convolutional networks.
\newblock In {\em Int. Conf. Comput. Vis.}, pages 764--773, 2017.

\bibitem{zhu2019dcnv2}
Xizhou Zhu, Han Hu, Stephen Lin, and Jifeng Dai.
\newblock Deformable convnets v2: More deformable, better results.
\newblock In {\em IEEE Conf. Comput. Vis. Pattern Recog.}, pages 9308--9316,
  2019.

\bibitem{liu2022slak}
Shiwei Liu, Tianlong Chen, Xiaohan Chen, Xuxi Chen, Qiao Xiao, Boqian Wu,
  Mykola Pechenizkiy, Decebal Mocanu, and Zhangyang Wang.
\newblock More convnets in the 2020s: Scaling up kernels beyond 51x51 using
  sparsity.
\newblock {\em arXiv preprint arXiv:2207.03620}, 2022.

\bibitem{zhai2022scaling}
Xiaohua Zhai, Alexander Kolesnikov, Neil Houlsby, and Lucas Beyer.
\newblock Scaling vision transformers.
\newblock In {\em IEEE Conf. Comput. Vis. Pattern Recog.}, pages 12104--12113,
  2022.

\bibitem{deng2009imagenet}
Jia Deng, Wei Dong, Richard Socher, Li-Jia Li, Kai Li, and Li~Fei-Fei.
\newblock Imagenet: A large-scale hierarchical image database.
\newblock In {\em IEEE Conf. Comput. Vis. Pattern Recog.}, pages 248--255,
  2009.

\bibitem{lin2014microsoft}
Tsung-Yi Lin, Michael Maire, Serge Belongie, James Hays, Pietro Perona, Deva
  Ramanan, Piotr Doll{\'a}r, and C~Lawrence Zitnick.
\newblock Microsoft coco: Common objects in context.
\newblock In {\em Eur. Conf. Comput. Vis.}, pages 740--755, 2014.

\bibitem{krizhevsky2017imagenet}
Alex Krizhevsky, Ilya Sutskever, and Geoffrey~E Hinton.
\newblock Imagenet classification with deep convolutional neural networks.
\newblock {\em Communications of the ACM}, 60(6):84--90, 2017.

\bibitem{simonyan2014very}
Karen Simonyan and Andrew Zisserman.
\newblock Very deep convolutional networks for large-scale image recognition.
\newblock {\em arXiv preprint arXiv:1409.1556}, 2014.

\bibitem{szegedy2015going}
Christian Szegedy, Wei Liu, Yangqing Jia, Pierre Sermanet, Scott Reed, Dragomir
  Anguelov, Dumitru Erhan, Vincent Vanhoucke, and Andrew Rabinovich.
\newblock Going deeper with convolutions.
\newblock In {\em IEEE Conf. Comput. Vis. Pattern Recog.}, pages 1--9, 2015.

\bibitem{he2016deep}
Kaiming He, Xiangyu Zhang, Shaoqing Ren, and Jian Sun.
\newblock Deep residual learning for image recognition.
\newblock In {\em IEEE Conf. Comput. Vis. Pattern Recog.}, pages 770--778,
  2016.

\bibitem{xie2017aggregated}
Saining Xie, Ross Girshick, Piotr Doll{\'a}r, Zhuowen Tu, and Kaiming He.
\newblock Aggregated residual transformations for deep neural networks.
\newblock In {\em IEEE Conf. Comput. Vis. Pattern Recog.}, pages 1492--1500,
  2017.

\bibitem{tan2019efficientnet}
Mingxing Tan and Quoc Le.
\newblock Efficientnet: Rethinking model scaling for convolutional neural
  networks.
\newblock In {\em International Conference on Machine Learning.}, pages
  6105--6114. PMLR, 2019.

\bibitem{tan2021efficientnetv2}
Mingxing Tan and Quoc Le.
\newblock Efficientnetv2: Smaller models and faster training.
\newblock In {\em International Conference on Machine Learning.}, pages
  10096--10106. PMLR, 2021.

\bibitem{howard2017mobilenets}
Andrew~G Howard, Menglong Zhu, Bo~Chen, Dmitry Kalenichenko, Weijun Wang,
  Tobias Weyand, Marco Andreetto, and Hartwig Adam.
\newblock Mobilenets: Efficient convolutional neural networks for mobile vision
  applications.
\newblock {\em arXiv preprint arXiv:1704.04861}, 2017.

\bibitem{ding2022scaling}
Xiaohan Ding, Xiangyu Zhang, Jungong Han, and Guiguang Ding.
\newblock Scaling up your kernels to 31x31: Revisiting large kernel design in
  cnns.
\newblock In {\em IEEE Conf. Comput. Vis. Pattern Recog.}, pages 11963--11975,
  2022.

\bibitem{liu2022more}
Shiwei Liu, Tianlong Chen, Xiaohan Chen, Xuxi Chen, Qiao Xiao, Boqian Wu,
  Mykola Pechenizkiy, Decebal Mocanu, and Zhangyang Wang.
\newblock More convnets in the 2020s: Scaling up kernels beyond 51x51 using
  sparsity.
\newblock {\em arXiv preprint arXiv:2207.03620}, 2022.

\bibitem{rao2022hornet}
Yongming Rao, Wenliang Zhao, Yansong Tang, Jie Zhou, Ser-Nam Lim, and Jiwen Lu.
\newblock Hornet: Efficient high-order spatial interactions with recursive
  gated convolutions.
\newblock {\em arXiv preprint arXiv:2207.14284}, 2022.

\bibitem{han2021connection}
Qi~Han, Zejia Fan, Qi~Dai, Lei Sun, Ming-Ming Cheng, Jiaying Liu, and Jingdong
  Wang.
\newblock On the connection between local attention and dynamic depth-wise
  convolution.
\newblock In {\em Int. Conf. Learn. Represent.}, 2021.

\bibitem{wang2020linformer}
Sinong Wang, Belinda~Z Li, Madian Khabsa, Han Fang, and Hao Ma.
\newblock Linformer: Self-attention with linear complexity.
\newblock {\em arXiv preprint arXiv:2006.04768}, 2020.

\bibitem{xia2022vision}
Zhuofan Xia, Xuran Pan, Shiji Song, Li~Erran Li, and Gao Huang.
\newblock Vision transformer with deformable attention.
\newblock In {\em IEEE Conf. Comput. Vis. Pattern Recog.}, pages 4794--4803,
  2022.

\bibitem{vaswani2021scaling}
Ashish Vaswani, Prajit Ramachandran, Aravind Srinivas, Niki Parmar, Blake
  Hechtman, and Jonathon Shlens.
\newblock Scaling local self-attention for parameter efficient visual
  backbones.
\newblock In {\em IEEE Conf. Comput. Vis. Pattern Recog.}, pages 12894--12904,
  2021.

\bibitem{kaplan2020scaling}
Jared Kaplan, Sam McCandlish, Tom Henighan, Tom~B Brown, Benjamin Chess, Rewon
  Child, Scott Gray, Alec Radford, Jeffrey Wu, and Dario Amodei.
\newblock Scaling laws for neural language models.
\newblock {\em arXiv preprint arXiv:2001.08361}, 2020.

\bibitem{ding2022davit}
Mingyu Ding, Bin Xiao, Noel Codella, Ping Luo, Jingdong Wang, and Lu~Yuan.
\newblock Davit: Dual attention vision transformers.
\newblock {\em arXiv preprint arXiv:2204.03645}, 2022.

\bibitem{zhu2019empirical}
Xizhou Zhu, Dazhi Cheng, Zheng Zhang, Stephen Lin, and Jifeng Dai.
\newblock An empirical study of spatial attention mechanisms in deep networks.
\newblock In {\em Int. Conf. Comput. Vis.}, pages 6688--6697, 2019.

\bibitem{chen2017deeplab}
Liang-Chieh Chen, George Papandreou, Iasonas Kokkinos, Kevin Murphy, and Alan~L
  Yuille.
\newblock Deeplab: Semantic image segmentation with deep convolutional nets,
  atrous convolution, and fully connected crfs.
\newblock {\em IEEE Trans. Pattern Anal. Mach. Intell.}, 40(4):834--848, 2017.

\bibitem{florian2017rethinking}
L-CCGP Florian and Schroff~Hartwig Adam.
\newblock Rethinking atrous convolution for semantic image segmentation.
\newblock In {\em IEEE Conf. Comput. Vis. Pattern Recog.}, volume~6, 2017.

\bibitem{chen2018encoder}
Liang-Chieh Chen, Yukun Zhu, George Papandreou, Florian Schroff, and Hartwig
  Adam.
\newblock Encoder-decoder with atrous separable convolution for semantic image
  segmentation.
\newblock In {\em Eur. Conf. Comput. Vis.}, pages 801--818, 2018.

\bibitem{lecun1989backpropagation}
Yann LeCun, Bernhard Boser, John~S Denker, Donnie Henderson, Richard~E Howard,
  Wayne Hubbard, and Lawrence~D Jackel.
\newblock Backpropagation applied to handwritten zip code recognition.
\newblock {\em Neural Computation}, 1(4):541--551, 1989.

\bibitem{chollet2017xception}
Fran{\c{c}}ois Chollet.
\newblock Xception: Deep learning with depthwise separable convolutions.
\newblock In {\em IEEE Conf. Comput. Vis. Pattern Recog.}, pages 1251--1258,
  2017.

\bibitem{zhu2020deformabledetr}
Xizhou Zhu, Weijie Su, Lewei Lu, Bin Li, Xiaogang Wang, and Jifeng Dai.
\newblock Deformable detr: Deformable transformers for end-to-end object
  detection.
\newblock {\em arXiv preprint arXiv:2010.04159}, 2020.

\bibitem{xiong2020layer}
Ruibin Xiong, Yunchang Yang, Di~He, Kai Zheng, Shuxin Zheng, Chen Xing,
  Huishuai Zhang, Yanyan Lan, Liwei Wang, and Tieyan Liu.
\newblock On layer normalization in the transformer architecture.
\newblock In {\em International Conference on Machine Learning.}, pages
  10524--10533. PMLR, 2020.

\bibitem{touvron2021training}
Hugo Touvron, Matthieu Cord, Matthijs Douze, Francisco Massa, Alexandre
  Sablayrolles, and Herv{\'e} J{\'e}gou.
\newblock Training data-efficient image transformers \& distillation through
  attention.
\newblock In {\em International Conference on Machine Learning.}, pages
  10347--10357, 2021.

\bibitem{yuan2021florence}
Lu~Yuan, Dongdong Chen, Yi-Ling Chen, Noel Codella, Xiyang Dai, Jianfeng Gao,
  Houdong Hu, Xuedong Huang, Boxin Li, Chunyuan Li, et~al.
\newblock Florence: A new foundation model for computer vision.
\newblock {\em arXiv preprint arXiv:2111.11432}, 2021.

\bibitem{su2022towards}
Weijie Su, Xizhou Zhu, Chenxin Tao, Lewei Lu, Bin Li, Gao Huang, Yu~Qiao,
  Xiaogang Wang, Jie Zhou, and Jifeng Dai.
\newblock Towards all-in-one pre-training via maximizing multi-modal mutual
  information.
\newblock {\em arXiv preprint arXiv:2211.09807}, 2022.

\bibitem{schuhmann2021laion}
Christoph Schuhmann, Richard Vencu, Romain Beaumont, Robert Kaczmarczyk,
  Clayton Mullis, Aarush Katta, Theo Coombes, Jenia Jitsev, and Aran
  Komatsuzaki.
\newblock Laion-400m: Open dataset of clip-filtered 400 million image-text
  pairs.
\newblock {\em arXiv preprint arXiv:2111.02114}, 2021.

\bibitem{thomee2016yfcc100m}
Bart Thomee, David~A Shamma, Gerald Friedland, Benjamin Elizalde, Karl Ni,
  Douglas Poland, Damian Borth, and Li-Jia Li.
\newblock Yfcc100m: The new data in multimedia research.
\newblock {\em Communications of the ACM}, 59(2):64--73, 2016.

\bibitem{changpinyo2021conceptual}
Soravit Changpinyo, Piyush Sharma, Nan Ding, and Radu Soricut.
\newblock Conceptual 12m: Pushing web-scale image-text pre-training to
  recognize long-tail visual concepts.
\newblock In {\em IEEE Conf. Comput. Vis. Pattern Recog.}, pages 3558--3568,
  2021.

\bibitem{touvron2022deit3}
Hugo Touvron, Matthieu Cord, and Herv{\'e} J{\'e}gou.
\newblock Deit iii: Revenge of the vit.
\newblock {\em arXiv preprint arXiv:2204.07118}, 2022.

\bibitem{yang2021focal}
Jianwei Yang, Chunyuan Li, Pengchuan Zhang, Xiyang Dai, Bin Xiao, Lu~Yuan, and
  Jianfeng Gao.
\newblock Focal self-attention for local-global interactions in vision
  transformers.
\newblock {\em arXiv preprint arXiv:2107.00641}, 2021.

\bibitem{hassani2021dilated}
Ismail~Khalfaoui Hassani, Thomas Pellegrini, and Timoth{\'e}e Masquelier.
\newblock Dilated convolution with learnable spacings.
\newblock {\em arXiv preprint arXiv:2112.03740}, 2021.

\bibitem{kolesnikov2020big}
Alexander Kolesnikov, Lucas Beyer, Xiaohua Zhai, Joan Puigcerver, Jessica Yung,
  Sylvain Gelly, and Neil Houlsby.
\newblock Big transfer (bit): General visual representation learning.
\newblock In {\em Eur. Conf. Comput. Vis.}, pages 491--507. Springer, 2020.

\bibitem{xie2020self}
Qizhe Xie, Minh-Thang Luong, Eduard Hovy, and Quoc~V Le.
\newblock Self-training with noisy student improves imagenet classification.
\newblock In {\em IEEE Conf. Comput. Vis. Pattern Recog.}, pages 10687--10698,
  2020.

\bibitem{chen2022vitadapter}
Zhe Chen, Yuchen Duan, Wenhai Wang, Junjun He, Tong Lu, Jifeng Dai, and
  Yu~Qiao.
\newblock Vision transformer adapter for dense predictions.
\newblock {\em arXiv preprint arXiv:2205.08534}, 2022.

\bibitem{he2017mask}
Kaiming He, Georgia Gkioxari, Piotr Doll{\'a}r, and Ross Girshick.
\newblock Mask r-cnn.
\newblock In {\em Int. Conf. Comput. Vis.}, pages 2961--2969, 2017.

\bibitem{cai2019cascade}
Zhaowei Cai and Nuno Vasconcelos.
\newblock Cascade r-cnn: high quality object detection and instance
  segmentation.
\newblock {\em IEEE Trans. Pattern Anal. Mach. Intell.}, 43(5):1483--1498,
  2019.

\bibitem{dai2021dynamic}
Xiyang Dai, Yinpeng Chen, Bin Xiao, Dongdong Chen, Mengchen Liu, Lu~Yuan, and
  Lei Zhang.
\newblock Dynamic head: Unifying object detection heads with attentions.
\newblock In {\em IEEE Conf. Comput. Vis. Pattern Recog.}, pages 7373--7382,
  2021.

\bibitem{xu2021end}
Mengde Xu, Zheng Zhang, Han Hu, Jianfeng Wang, Lijuan Wang, Fangyun Wei, Xiang
  Bai, and Zicheng Liu.
\newblock End-to-end semi-supervised object detection with soft teacher.
\newblock In {\em Int. Conf. Comput. Vis.}, pages 3060--3069, 2021.

\bibitem{zhang2022dino}
Hao Zhang, Feng Li, Shilong Liu, Lei Zhang, Hang Su, Jun Zhu, Lionel~M Ni, and
  Heung-Yeung Shum.
\newblock Dino: Detr with improved denoising anchor boxes for end-to-end object
  detection.
\newblock {\em arXiv preprint arXiv:2203.03605}, 2022.

\bibitem{yang2022focal}
Jianwei Yang, Chunyuan Li, and Jianfeng Gao.
\newblock Focal modulation networks.
\newblock {\em arXiv preprint arXiv:2203.11926}, 2022.

\bibitem{chen2022group}
Qiang Chen, Xiaokang Chen, Gang Zeng, and Jingdong Wang.
\newblock Group detr: Fast training convergence with decoupled one-to-many
  label assignment.
\newblock {\em arXiv preprint arXiv:2207.13085}, 2022.

\bibitem{li2022exploring}
Yanghao Li, Hanzi Mao, Ross Girshick, and Kaiming He.
\newblock Exploring plain vision transformer backbones for object detection.
\newblock {\em arXiv preprint arXiv:2203.16527}, 2022.

\bibitem{liang2022cbnet}
Tingting Liang, Xiaojie Chu, Yudong Liu, Yongtao Wang, Zhi Tang, Wei Chu,
  Jingdong Chen, and Haibin Ling.
\newblock Cbnet: A composite backbone network architecture for object
  detection.
\newblock {\em IEEE Trans. Image Process.}, 2022.

\bibitem{shao2019objects365}
Shuai Shao, Zeming Li, Tianyuan Zhang, Chao Peng, Gang Yu, Xiangyu Zhang, Jing
  Li, and Jian Sun.
\newblock Objects365: A large-scale, high-quality dataset for object detection.
\newblock In {\em Int. Conf. Comput. Vis.}, pages 8430--8439, 2019.

\bibitem{cheng2021masked}
Bowen Cheng, Ishan Misra, Alexander~G Schwing, Alexander Kirillov, and Rohit
  Girdhar.
\newblock Masked-attention mask transformer for universal image segmentation.
\newblock {\em arXiv preprint arXiv:2112.01527}, 2021.

\bibitem{xiao2018unified}
Tete Xiao, Yingcheng Liu, Bolei Zhou, Yuning Jiang, and Jian Sun.
\newblock Unified perceptual parsing for scene understanding.
\newblock In {\em Eur. Conf. Comput. Vis.}, pages 418--434, 2018.

\bibitem{zhou2017scene}
Bolei Zhou, Hang Zhao, Xavier Puig, Sanja Fidler, Adela Barriuso, and Antonio
  Torralba.
\newblock Scene parsing through ade20k dataset.
\newblock In {\em IEEE Conf. Comput. Vis. Pattern Recog.}, pages 633--641,
  2017.

\bibitem{xie2021segformer}
Enze Xie, Wenhai Wang, Zhiding Yu, Anima Anandkumar, Jose~M Alvarez, and Ping
  Luo.
\newblock Segformer: Simple and efficient design for semantic segmentation with
  transformers.
\newblock {\em Adv. Neural Inform. Process. Syst.}, 34, 2021.

\bibitem{loshchilov2017decoupled}
Ilya Loshchilov and Frank Hutter.
\newblock Decoupled weight decay regularization.
\newblock {\em arXiv preprint arXiv:1711.05101}, 2017.

\bibitem{caesar2018coco}
Holger Caesar, Jasper Uijlings, and Vittorio Ferrari.
\newblock Coco-stuff: Thing and stuff classes in context.
\newblock In {\em IEEE Conf. Comput. Vis. Pattern Recog.}, pages 1209--1218,
  2018.

\bibitem{diao2022metaformer}
Qishuai Diao, Yi~Jiang, Bin Wen, Jia Sun, and Zehuan Yuan.
\newblock Metaformer: A unified meta framework for fine-grained recognition.
\newblock {\em arXiv preprint arXiv:2203.02751}, 2022.

\bibitem{liu2022mixmim}
Jihao Liu, Xin Huang, Yu~Liu, and Hongsheng Li.
\newblock Mixmim: Mixed and masked image modeling for efficient visual
  representation learning.
\newblock {\em arXiv preprint arXiv:2205.13137}, 2022.

\bibitem{singh2022revisiting}
Mannat Singh, Laura Gustafson, Aaron Adcock, Vinicius de~Freitas~Reis, Bugra
  Gedik, Raj~Prateek Kosaraju, Dhruv Mahajan, Ross Girshick, Piotr Doll{\'a}r,
  and Laurens Van Der~Maaten.
\newblock Revisiting weakly supervised pre-training of visual perception
  models.
\newblock In {\em IEEE Conf. Comput. Vis. Pattern Recog.}, pages 804--814,
  2022.

\bibitem{liu2021polarized}
Huajun Liu, Fuqiang Liu, Xinyi Fan, and Dong Huang.
\newblock Polarized self-attention: Towards high-quality pixel-wise regression.
\newblock {\em arXiv preprint arXiv:2107.00782}, 2021.

\bibitem{liu2022cmx}
Huayao Liu, Jiaming Zhang, Kailun Yang, Xinxin Hu, and Rainer Stiefelhagen.
\newblock Cmx: Cross-modal fusion for rgb-x semantic segmentation with
  transformers.
\newblock {\em arXiv preprint arXiv:2203.04838}, 2022.

\bibitem{zhang2022glipv2}
Haotian Zhang, Pengchuan Zhang, Xiaowei Hu, Yen-Chun Chen, Liunian~Harold Li,
  Xiyang Dai, Lijuan Wang, Lu~Yuan, Jenq-Neng Hwang, and Jianfeng Gao.
\newblock Glipv2: Unifying localization and vision-language understanding.
\newblock {\em arXiv preprint arXiv:2206.05836}, 2022.

\bibitem{fang2022eva}
Yuxin Fang, Wen Wang, Binhui Xie, Quan Sun, Ledell Wu, Xinggang Wang, Tiejun
  Huang, Xinlong Wang, and Yue Cao.
\newblock Eva: Exploring the limits of masked visual representation learning at
  scale.
\newblock {\em arXiv preprint arXiv:2211.07636}, 2022.

\bibitem{ghiasi2021simple}
Golnaz Ghiasi, Yin Cui, Aravind Srinivas, Rui Qian, Tsung-Yi Lin, Ekin~D Cubuk,
  Quoc~V Le, and Barret Zoph.
\newblock Simple copy-paste is a strong data augmentation method for instance
  segmentation.
\newblock In {\em IEEE Conf. Comput. Vis. Pattern Recog.}, pages 2918--2928,
  2021.

\bibitem{jin2016ATLDETv2}
Xuan Jin, Wei Su, Rong Zhang, Yuan He, and Hui Xue.
\newblock Atldetv2.
\newblock
  \url{http://host.robots.ox.ac.uk/leaderboard/displaylb_main.php?challengeid=11&compid=4},
  2019.

\bibitem{liu20201st}
Yu~Liu, Guanglu Song, Yuhang Zang, Yan Gao, Enze Xie, Junjie Yan, Chen~Change
  Loy, and Xiaogang Wang.
\newblock 1st place solutions for openimage2019--object detection and instance
  segmentation.
\newblock {\em arXiv preprint arXiv:2003.07557}, 2020.

\bibitem{zheng2022progressive}
Anlin Zheng, Yuang Zhang, Xiangyu Zhang, Xiaojuan Qi, and Jian Sun.
\newblock Progressive end-to-end object detection in crowded scenes.
\newblock In {\em IEEE Conf. Comput. Vis. Pattern Recog.}, pages 857--866,
  2022.

\bibitem{xu2022pp}
Shangliang Xu, Xinxin Wang, Wenyu Lv, Qinyao Chang, Cheng Cui, Kaipeng Deng,
  Guanzhong Wang, Qingqing Dang, Shengyu Wei, Yuning Du, et~al.
\newblock Pp-yoloe: An evolved version of yolo.
\newblock {\em arXiv preprint arXiv:2203.16250}, 2022.

\bibitem{van2018inaturalist}
Grant Van~Horn, Oisin Mac~Aodha, Yang Song, Yin Cui, Chen Sun, Alex Shepard,
  Hartwig Adam, Pietro Perona, and Serge Belongie.
\newblock The inaturalist species classification and detection dataset.
\newblock In {\em IEEE Conf. Comput. Vis. Pattern Recog.}, pages 8769--8778,
  2018.

\bibitem{zhou2014learning}
Bolei Zhou, Agata Lapedriza, Jianxiong Xiao, Antonio Torralba, and Aude Oliva.
\newblock Learning deep features for scene recognition using places database.
\newblock {\em Adv. Neural Inform. Process. Syst.}, 27, 2014.

\bibitem{lopez2020semantic}
Alejandro L{\'o}pez-Cifuentes, Marcos Escudero-Vinolo, Jes{\'u}s Besc{\'o}s,
  and {\'A}lvaro Garc{\'\i}a-Mart{\'\i}n.
\newblock Semantic-aware scene recognition.
\newblock {\em Pattern Recognition}, 102:107256, 2020.

\bibitem{gupta2019lvis}
Agrim Gupta, Piotr Dollar, and Ross Girshick.
\newblock Lvis: A dataset for large vocabulary instance segmentation.
\newblock In {\em IEEE Conf. Comput. Vis. Pattern Recog.}, pages 5356--5364,
  2019.

\bibitem{everingham2010pascal}
Mark Everingham, Luc Van~Gool, Christopher~KI Williams, John Winn, and Andrew
  Zisserman.
\newblock The pascal visual object classes (voc) challenge.
\newblock {\em Int. J. Comput. Vis.}, 88(2):303--338, 2010.

\bibitem{kuznetsova2020open}
Alina Kuznetsova, Hassan Rom, Neil Alldrin, Jasper Uijlings, Ivan Krasin, Jordi
  Pont-Tuset, Shahab Kamali, Stefan Popov, Matteo Malloci, Alexander
  Kolesnikov, et~al.
\newblock The open images dataset v4.
\newblock {\em Int. J. Comput. Vis.}, 128(7):1956--1981, 2020.

\bibitem{shao2018crowdhuman}
Shuai Shao, Zijian Zhao, Boxun Li, Tete Xiao, Gang Yu, Xiangyu Zhang, and Jian
  Sun.
\newblock Crowdhuman: A benchmark for detecting human in a crowd.
\newblock {\em arXiv preprint arXiv:1805.00123}, 2018.

\bibitem{yu2020bdd100k}
Fisher Yu, Haofeng Chen, Xin Wang, Wenqi Xian, Yingying Chen, Fangchen Liu,
  Vashisht Madhavan, and Trevor Darrell.
\newblock Bdd100k: A diverse driving dataset for heterogeneous multitask
  learning.
\newblock In {\em IEEE Conf. Comput. Vis. Pattern Recog.}, pages 2636--2645,
  2020.

\bibitem{mottaghi2014role}
Roozbeh Mottaghi, Xianjie Chen, Xiaobai Liu, Nam-Gyu Cho, Seong-Whan Lee, Sanja
  Fidler, Raquel Urtasun, and Alan Yuille.
\newblock The role of context for object detection and semantic segmentation in
  the wild.
\newblock In {\em IEEE Conf. Comput. Vis. Pattern Recog.}, pages 891--898,
  2014.

\bibitem{Cordts_2016_CVPR}
Marius Cordts, Mohamed Omran, Sebastian Ramos, Timo Rehfeld, Markus Enzweiler,
  Rodrigo Benenson, Uwe Franke, Stefan Roth, and Bernt Schiele.
\newblock The cityscapes dataset for semantic urban scene understanding.
\newblock In {\em IEEE Conf. Comput. Vis. Pattern Recog.}, 2016.

\bibitem{tao2020hierarchical}
Andrew Tao, Karan Sapra, and Bryan Catanzaro.
\newblock Hierarchical multi-scale attention for semantic segmentation.
\newblock {\em arXiv preprint arXiv:2005.10821}, 2020.

\bibitem{neuhold2017mapillary}
Gerhard Neuhold, Tobias Ollmann, Samuel Rota~Bulo, and Peter Kontschieder.
\newblock The mapillary vistas dataset for semantic understanding of street
  scenes.
\newblock In {\em Int. Conf. Comput. Vis.}, pages 4990--4999, 2017.

\bibitem{silberman2012indoor}
Nathan Silberman, Derek Hoiem, Pushmeet Kohli, and Rob Fergus.
\newblock Indoor segmentation and support inference from rgbd images.
\newblock {\em Eur. Conf. Comput. Vis.}, 7576:746--760, 2012.

\end{thebibliography}
}

\end{document}